\documentclass[conference]{IEEEtran}
\IEEEoverridecommandlockouts
\usepackage{cite}
\usepackage{amsmath,amssymb,amsfonts}
\usepackage{algorithmic}
\usepackage{graphicx}
\usepackage{textcomp}
\usepackage{xcolor}
\usepackage{authblk}
\usepackage{subfigure}

\def\BibTeX{{\rm B\kern-.05em{\sc i\kern-.025em b}\kern-.08em
    T\kern-.1667em\lower.7ex\hbox{E}\kern-.125emX}}
\begin{document}

\title{A study of traits that affect learnability in GANs\\
}

\author[1,2]{Niladri Shekhar Dutt}
\author[2]{Sunil Patel}
\affil[1]{SRM Institute of Science and Technology, Chennai, India \authorcr nd9763@srmist.edu.in \vspace{1.5ex}}
\affil[2]{Nvidia \authorcr supatel@nvidia.com \vspace{-2ex}}

\maketitle

\begin{abstract}
Generative Adversarial Networks (GANs) are algorithmic architectures that use two neural networks, pitting one against the opposite (thus the “adversarial”) so as to come up with new, synthetic instances of data that can pass for real data. Training a GAN is a challenging problem which requires us to apply advanced techniques like hyperparameter tuning, architecture engineering etc. Many different losses, regularization and normalisation schemes, network architectures have been proposed to solve this challenging problem for different types of datasets. It becomes necessary to understand the experimental observations and deduce a simple theory for it. In this paper, we perform empirical experiments using parameterized synthetic datasets to probe what traits affect learnability.
\end{abstract}


\section{Introduction}
Generative Adversarial Networks (GAN) are a type generative models which are used for various use cases like image editing,  image generation, security and strengthening existing models, semi-supervised learning, and attention prediction. \cite{radford2015unsupervised}\cite{zhang2017stackgan}. In the GAN framework, the model tries to match a simple distribution with the data distribution by learning a deterministic transformation. During the learning process, the generator learns to generate images which are almost similar to the real data. However, the discriminator learns to differentiate between the real data and the generated data. This becomes a game-like scenario where the solution to the game is the nash equilibrium where the players can't improve their cost unilaterally. \cite{goodfellow2014generative}.
There are quite a few flavours of GAN that have been proposed which are both unsupervised as well as conditional. Although these algorithms are able to achieve great results, it is still not clear which algorithms perform better on a particular dataset. Limited comparisons and a lack of robust and consistent metrics make it difficult to compare the algorithms. The computational budget to search all the hyperparameters is also a restricting parameter. 

The main motive of this study is to help to choose a better and more relevant model, according to the dataset in hand, from the myriad of options available. It will also help us in getting a better understanding of the model and make relevant modifications which are critical for better results. One cannot explicitly compute the probability of pg(x) which becomes a bottleneck. Therefore, log-likelihood is not preferably used on test sets as evaluations cannot be conducted \cite{lucic2018gans}.

So how do we understand which kind of distributions can GANs model? To answer this, We perform empirical experiments using parameterized synthetic datasets to identify and examine traits and characteristics that affect learnability. We choose synthetic dataset for our study since synthetic datasets can be parameterized according to desired needs like connectedness or smoothness. By identifying these traits we can particularly tackle key issues affecting learnability and prepare and process data such that the training landscape becomes smoother and more stable, and thus yielding better quality samples. This can be extremely useful to perform better architecture search especially in Generative Teaching Networks (GTNs) \cite{such2019generative} which generate synthetic data that can help other neural networks to rapidly learn the distribution when trained on real data.

\section{Literature Survey}

\subsection{Challenges in Generative Adversarial Networks}

There are several ongoing challenges in the study of GANs. Some challenges include mode collapse, failure to converge, and vanishing gradient. When the generator produces a limited diversity of samples irrespective of the sample, which is acceptable by the discriminator, it will always produce the same result and hence the training will be stuck (at local minima). GANs also frequently fail to converge as two neural networks competing against each other is a very unstable stable state. 

Another challenge is that when the generator doesn’t get enough information from the discriminator, during backpropagation, the gradient becomes increasingly smaller and the training stucks. Another challenge quantitatively evaluating the trained model. Model likelihood is the general approach to evaluate generative models. When considering complex high dimensional data, even log-likelihood approximation becomes extremely challenging. Yuhuai Wu et al \cite{wu2016quantitative} suggest an annealed importance sampling algorithm to estimate the hold-out log-likelihood. Assumption of the Gaussian observation model is major downside of this approach. Lucas Theis et al \cite{theis2015note} provide an analysis of common failure modes and demonstrate that it is possible to achieve high likelihood, but low visual quality, and vice-versa.  \cite{lucic2018gans}.

Ferenc Huszár \cite{huszar2015not} discusses the drawbacks of likelihood-based training and evaluation and hence ranking models based on these estimates is not recommended.


\subsection{Loss Functions}
Goodfellow \cite{goodfellow2014bengio} suggests two loss functions: the minimax GAN and the NS GAN. For binary classification task, minimax loss function minimizes the negative log likelihood. In NS GAN, the generator tries to maximize the probablity of generated samples being real. 
\begin{equation}
L_{Disc}= -E_{z \sim P}[\log{(1-Disc(z))}]- E_{\hat{z} \sim Q}[\log{(1-D(\hat{z}))}]
\label{eq:mkceqn}
\end{equation} 
\begin{equation}
L_{Gen}= -E_{\hat{z} \sim Q}[\log{(D(\hat{z}))}]
\label{eq:mkceqn}
\end{equation} 
where \(D(z)\) is the probability of \(z\) being sampled from \(D(z)\).
\\Next we consider WGAN \cite{pmlr-v70-arjovsky17a} where the author considers the wasserstein distance to calculate the loss as shown below. 
\begin{equation}
L_{Disc}= -E_{z \sim P}[D(z)]- E_{\hat{z} \sim Q}[D(\hat{z})]
\label{eq:mkceqn}
\end{equation} 
\begin{equation}
L_{Gen}= -E_{\hat{z} \sim Q}[Disc(\hat{z})]
\label{eq:mkceqn}
\end{equation} 
where the discriminator output \(Disc(z)\in R\) and \(Disc\) is required to be 1-Lipschitz. WGANs make the training process more stable. It also helps in making the training process less sensitive to model architecture.  \\ 
Next, we consider the least-square loss where we minimize Pearson \(\chi^2\) divergence between P and Q \cite{mao2017least}. The loss function is as follow:
\begin{equation}
L_{Disc}= -E_{z \sim P}[{(Disc(z-1))}^2]- E_{\hat{z} \sim Q}[{(Disc(\hat{z}))}^2]
\label{eq:mkceqn}
\end{equation} 
\begin{equation}
L_{Gen}= -E_{\hat{z} \sim Q}[{(Disc(\hat{z})-1)}^2]
\label{eq:mkceqn}
\end{equation} 
where \(Disc(z)\in R\) is the output of the discriminator. This smooth loss function saturates slower than the cross-entropy loss.


\subsection{DCGAN architecture}

We use a DCGAN \cite{radford2015unsupervised} to train and generate samples from the synthetic dataset. It mainly consists of convolution layers and for down sampling and up sampling it uses transposed convolution and convolutional stride. However, they do not contain fully connected layers or max pooling. The figure below is the network design for the generator. 

\begin{figure}[htpb]
\centering
\includegraphics[scale=0.5]{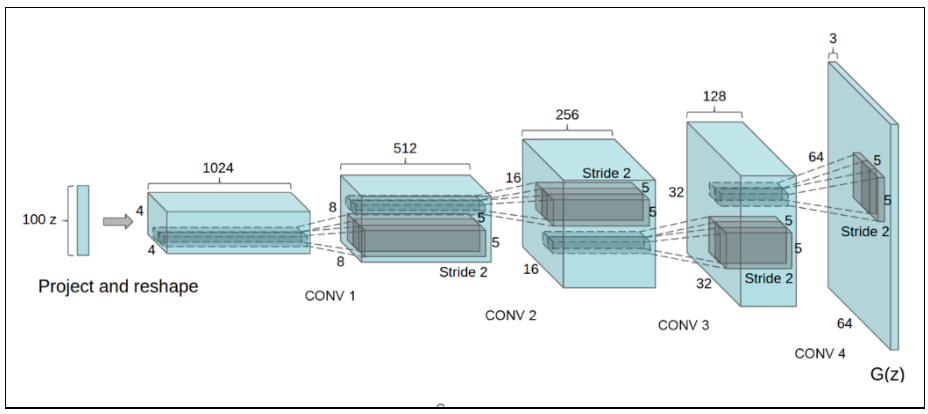}
\caption{\bf DCGAN generator presented in the LSUN scene modeling paper}
\label{fig:beam}
\end{figure}

In DCGAN, all the max pooling layers are replaced by convolutional stride. LeakyReLU is employed within the discriminator and ReLU is employed within the generator. The output layer uses tanh. Batch normalization is used in this model . As a result the Generator network is able to take random noise and map it into images. The resultant images are such that the discriminator cannot tell which images came from the dataset and which ones came from the generator.

\subsection{Comparing Models}
After deciding the loss functions, tuning the hyperparameters, determining the initialization method and training the model, the question of how to compare the models arises. In this study, we consider the criteria of inception score

\subsubsection{Inception Score}
According to Salimans et al.\cite{salimans2016improved}, Inception score can be found by calculating how much each image's label distribution differs from the marginal label distribution for the whole set of images. To understand the dependence of the regularity of the generated n-gons with the inception score, we varied the some properties of the synthetic dataset. We calculated the value of IC by varying the values of:
\begin{itemize}
\item Number of side of the n-gons.
\item Shifting to mean.
\item Minimum segment angle.
\end{itemize}
The authors found that this score is well-correlated with scores from human annotators.
\begin{equation}
Inception Score (Gen) = exp(E_{x \sim P_a}[D_{KL}[\:p(\:y\:|\:x\:)\:||\:p(\:y\:)]\:]
\label{eq:mkceqn}
\end{equation} 
If the GAN generates one image per class, then it can easily misrepresent the performance of the GAN. Also insensitivity to the prior distribution is a drawback for this method.
Some limitations of IC score are:
\begin{itemize}
    \item The generating image should be present in the classifiers training data, otherwise you may always get low IC value.
    \item if the generated labels are different than the training set, then ic may be low.
    \item if generator generates one image per class, repeating each image many times, it can score highly
\end{itemize}


\section{Synthetic Dataset Generation}
In order to understand the behaviour of the GANs that we are training, we generated 18 different synthetic dataset by varying number of vertices, minimum segment angle and shifting the data to mean. To get a better understanding and control over the training process, we generate our own dataset. We generated 18 datasets with 10,000 images each of convex polygons.\cite{chalmeta2013measuring} 
\begin{figure}[htpb]
\centering
\subfigure[Triangles]{
\includegraphics[scale=0.08]{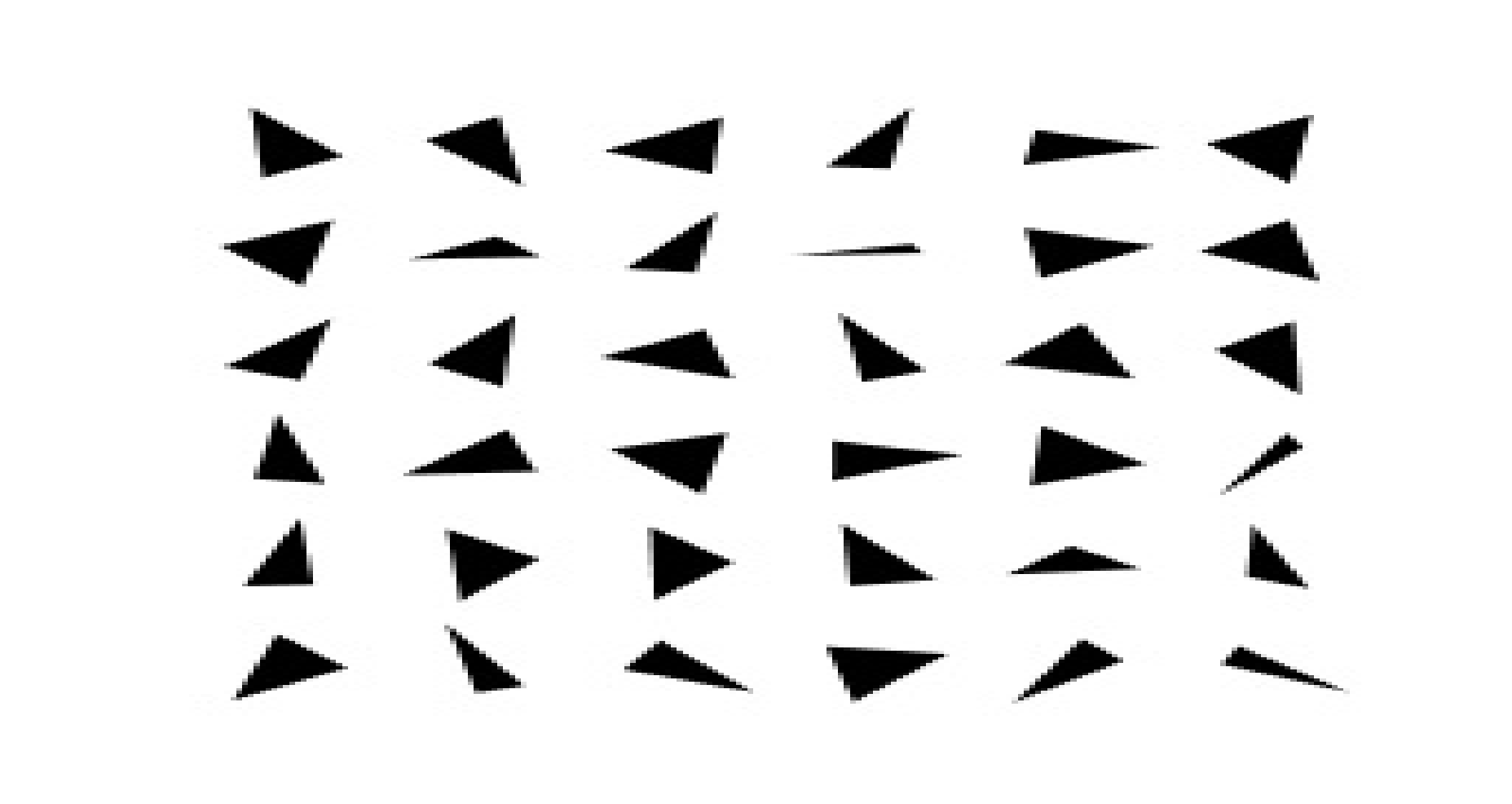}
\label{fig:ellipse}} 
\subfigure[Quadilateral]{
\includegraphics[scale=0.08]{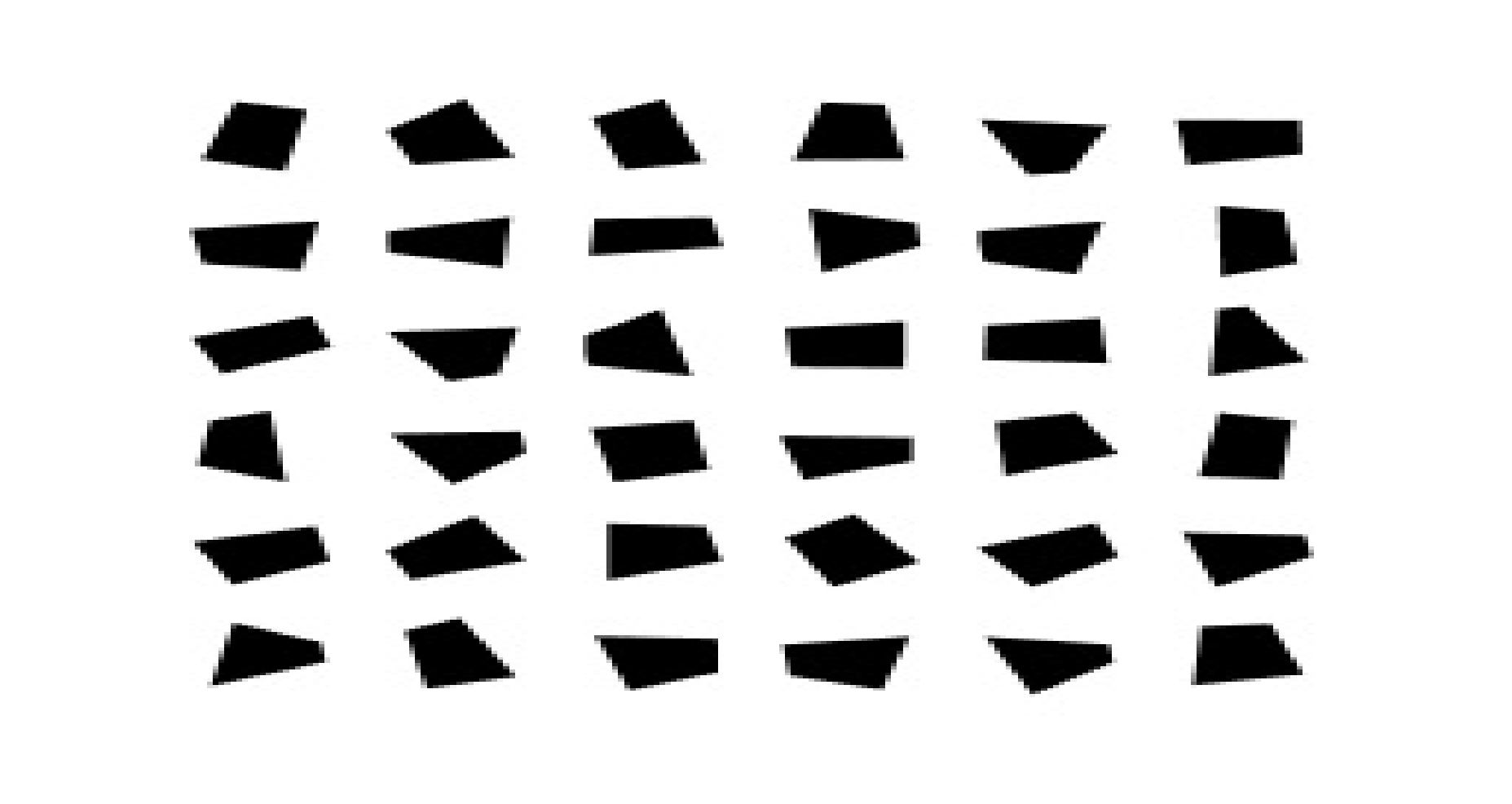}
\label{fig:scotch}}
\subfigure[Pentagon]{
\includegraphics[scale=0.08]{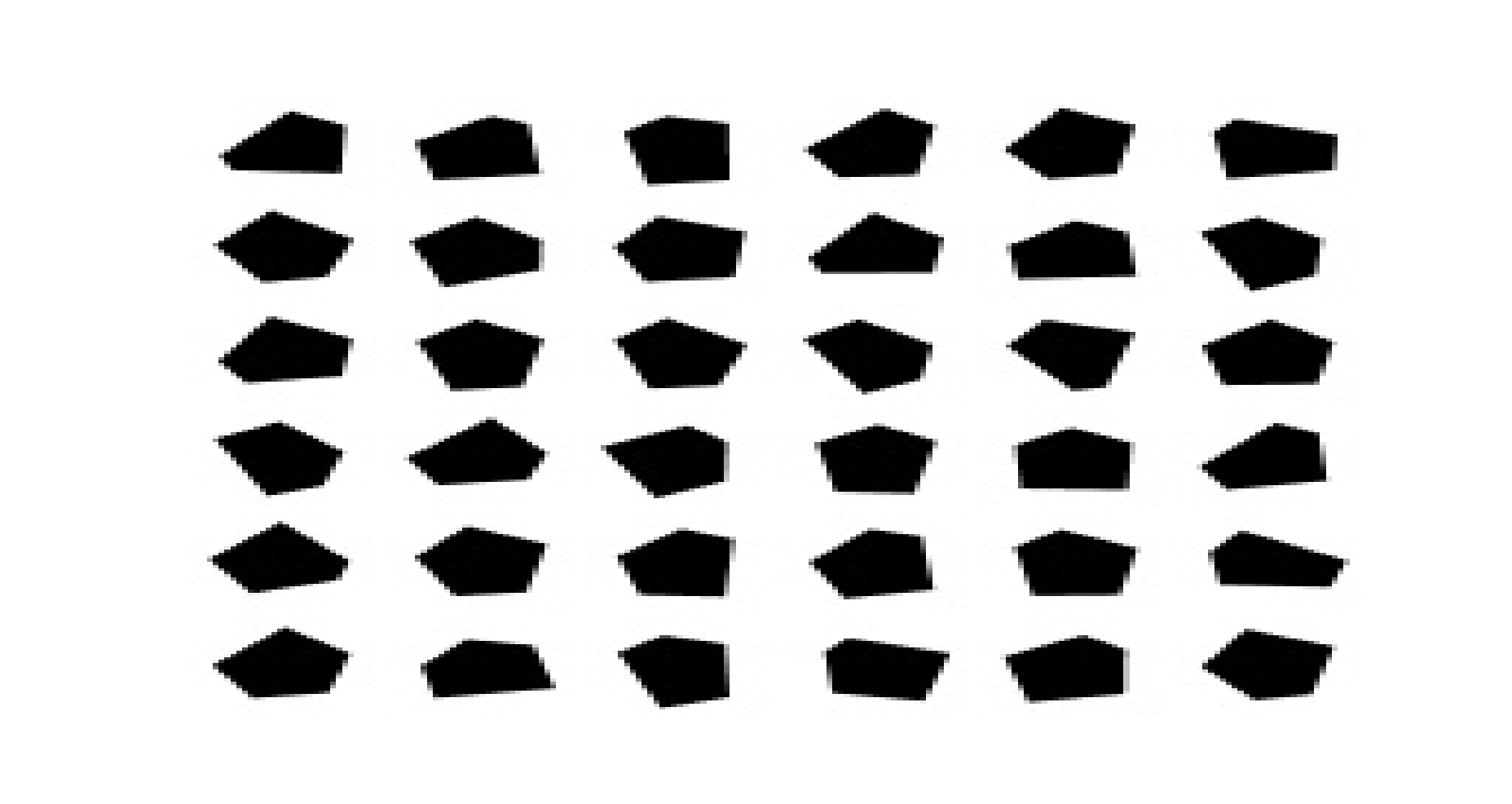}
\label{fig:scotch}}
\caption{\bf Synthetic Dataset}
\label{fig:4s}
\end{figure}

\subsection{Regularity of convex polygons}
In a convex polygon, all the interior angles are less than 180 degrees. Hence all the vertices of the polygon are pointing outwards, away from the interior of the shape. If we try to draw a line across a convex polygon, it will cut it exactly two times. For a polygon to be regular, there are a range of different aspects of regularity which we have considered while generating the synthetic dataset. The characteristics of the regularity for the generated samples depend on equality of angles and edge lengths, angular, optimal area-perimeter ratio, angular and areal symmetry. 
The following are the preferred properties for n-gon to be regular:
\begin{itemize}
\item The ratio of area to perimeter should be maximum.
\item The angles are similar.
\item The length of the sides are similar.
\item shift to mean
\item segment angle 
\end{itemize}
For n-gons to be regular, the ratio of area to perimeter should be maximum.

\subsection{Method to generate images}

To generate the images, we first take an ellipse and then divide the ellipse into n number of segments, where n is the number of vertices required for that polygon. Then a vertex is selected in each of these segments such that the minimum segment angle between vertex A, zero and vertex B is respected. This generates a different random convex polygon each time. The result can be seen in the diagram below.


\section{GAN architecture}

We use a DCGAN \cite{radford2015unsupervised} to train and generate samples from the synthetic dataset. It mainly consists of convolution layers without max pooling or fully connected layers. It uses convolutional stride and transposed convolution for the downsampling and the upsampling. The figure below is the network design for the generator.

In DCGAN, all the max pooling layers are replaced by convolutional stride. LeakyReLU is employed within the discriminator and ReLU is employed within the generator except for the output which uses tanh. Batch normalization is used in this model  except for the output layer for the generator and the input layer of the discriminator. As a result the Generator network is able to take random noise and map it into images. The resultant images are such that the discriminator cannot tell which images came from the dataset and which images came from the generator.

\subsection{Methodology}

By varying the parameters we generated multiple synthetic datasets with varying statistical characteristics such as number of vertices, minimum segment angle and shift towards mean. We then used DCGAN architecture to train on every generated dataset to see the quality and diversity of the generated samples, discriminator and adversarial loss curves, and stability of the GAN while training. We then used Inception Score (IS) to evaluate the quality of the generated samples as well as the quality of the generated synthetic dataset. We change three parameters while probing the experiment.
\begin{itemize}
    \item Number of vertices of the polygon
    \item Segment angle
    \item Shifting the data towards mean
\end{itemize}

\subsection{Shifting the data towards mean}
Shift to mean is the process of subtracting the values by the mean of the data that we are dealing with. Shifting to mean does not change the spread of the data, i.e. the range, IQR and the standard deviation will remain constant. The scaling is done by dividing the standard deviation. If a set of samples are both shifted to mean and divided by the standard deviation, the sample set is called normalized. In normalised data, the shape of the distribution does not change.

\subsection{Segment angle}
Segment angle, as the name suggest is the angle between two adjacent line segments in a polygon. It defines the shape and the characteristics of the polygon. The lower the segment angle, the higher the IS value and hence the better the GAN model.

\section{Generated shapes}
\begin{figure}[htpb]
\centering
\subfigure[Generated triangles]{
\includegraphics[scale=0.2]{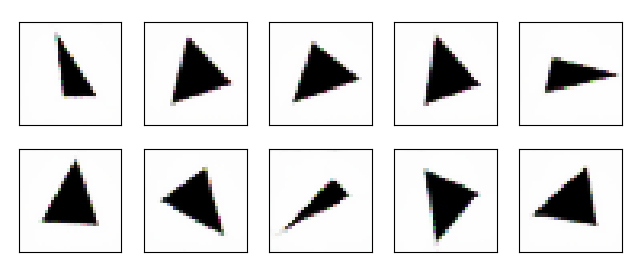}
\label{fig:ellipse}} 
\subfigure[Generated quadilaterals]{
\includegraphics[scale=0.2]{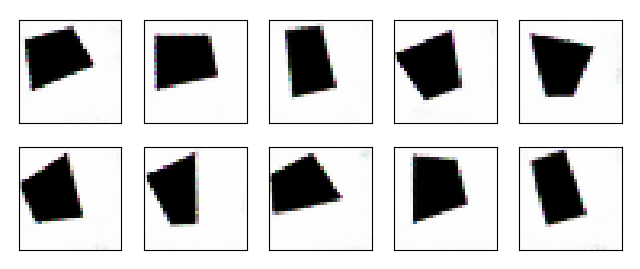}
\label{fig:scotch}}
\subfigure[Generated pentagons]{
\includegraphics[scale=0.2]{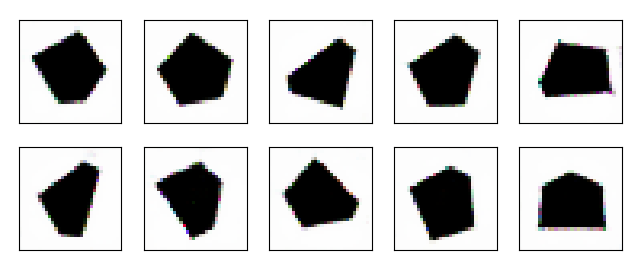}
\label{fig:scotch}}
\subfigure[Generated samples when all shapes are trained together]{
\includegraphics[scale=0.2]{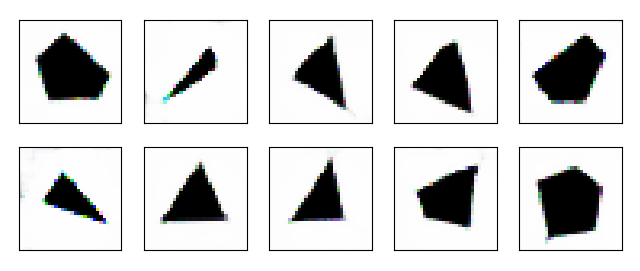}
\label{fig:scotch}}
\caption{Generated samples from synthetic data shifted to mean and minimum segment angle as 20}
\label{fig:gensam}
\end{figure}
After training the GAN models using the synthetic dataset, we obtained the following results for different shapes. During the training, we shifted to mean for the synthetic data and the minimum segment angle was set to be 20. Figure \ref{fig:gensam} shows the different generated samples for triangles, quadrilaterals, pentagons and a combined result.

\section{Calculating Inception Score.}

Inception score is found by calculating how much the each image's label distribution differs from the marginal label distribution for the whole set of images. To understand the dependence of the regularity of the generated n-gons with the inception score, we varied the some properties of the synthetic dataset like varying the values of number of side of the n-gons, shifting the data to mean and changing the minimum segment angle.
\begin{table}[htpb]
\centering
\caption{Synthetic Data}\label{tab: }
\begin{tabular}{|c|c|c|c|c|c|}
\hline
Vertices & Mean shift & Min segment angle & $IC_{avg}$ & $IC_{std}$ \\
\hline
3  &true &20  &1.124884 &0.003417 \\
3  &false &20  &1.128503 &0.003123 \\
3  &true &40  &1.111411 &0.002541 \\
3  &false &40  &1.124031 &0.003362 \\
3  &true &60  &1.100721 &0.001812 \\
3  &false &60  &1.114217 &0.002742 \\
4  &true &20  &1.190351 &0.004603 \\
4  &false &20  &1.171300 &0.006284 \\
4  &true &40  &1.163188 &0.005843 \\
4  &false &40  &1.163161 &0.006330 \\
4  &true &60  &1.148575 &0.003179 \\
4  &false &60  &1.148610 &0.003674 \\
5  &true &20  &1.103540 &0.003587 \\
5  &false &20  &1.136112 &0.003851 \\
5  &true &40  &1.091653 &0.002516 \\
5  &false &40  &1.122759 &0.002489 \\
5  &true &60  &1.100455 &0.003024 \\
5  &false &60  &1.100403 &0.002216 \\
\hline
\end{tabular}
\end{table}

\begin{table}[htpb]
\centering
\caption{Generated Samples}\label{tab: }
\begin{tabular}{|c|c|c|c|c|c|}
\hline
Vertices & Mean shift & Min segment angle & $IC_{avg}$ & $IC_{std}$ \\
\hline
3  &true &20  &1.0952661 &0.0378311 \\
3  &false &20  &1.1196281 &0.0160307 \\
3  &true &40  &1.0896549 &0.0190237 \\
3  &false &40  &1.1139643 &0.0312837 \\
3  &true &60  &1.1102858 &0.0456649 \\
3  &false &60  &1.1114547 &0.0251433 \\
4  &true &20  &1.1591276 &0.0369158 \\
4  &false &20  &1.1559708 &0.0677945 \\
4  &true &40  &1.1426497 &0.0562124 \\
4  &false &40  &1.1542552 &0.0470865 \\
4  &true &60  &1.1284901 &0.0439606 \\
4  &false &60  &1.1463740 &0.0406257 \\
5  &true &20  &1.0805178 &0.0160529 \\
5  &false &20  &1.1207236 &0.0322686 \\
5  &true &40  &1.0821022 &0.0280495 \\
5  &false &40  &1.1080600 &0.0291524 \\
5  &true &60  &1.0877767 &0.0144126 \\
5  &false &60  &1.0999079 &0.0246287 \\
\hline
\end{tabular}
\end{table}

We calculated the IS scores for a dataset which consisted of all types of n-gons, i.e. triangles, quadrilaterals and pentagons. The  table \ref{tab:a } and \ref{tab:b } show the result of the generated IS average values and the standard deviation.

We calculated the average values as well as the standard deviation in order to get a better understanding of the result and the dependence of regularity of generated dataset.

\begin{table}[htpb]
\centering
\caption{Synthetic Data}\label{tab:a }
\begin{tabular}{|c|c|c|c|c|c|}
\hline
Vertices & Mean shift & Min seg angle & $IC_{avg}$ & $IC_{std}$ \\
\hline
combined &true &20 &1.1897008 &0.006073 \\
combined &true &40 &1.1389768 &0.003560 \\
combined &true &60 &1.2094663 &0.003435 \\
\hline
\end{tabular}
\end{table}
\begin{table}[htpb]
\centering
\caption{Generated samples}\label{tab:b }
\begin{tabular}{|c|c|c|c|c|c|}
\hline
Vertices & Mean shift & Min seg angle & $IC_{avg}$ & $IC_{std}$ \\
\hline
combined &true &20 &1.2319351 &0.065050 \\
combined &true &40 &1.1263522 &0.059703 \\
combined &true &60 &1.2059145 &0.053481 \\
\hline
\end{tabular}
\end{table}

\section{Results Discussion}
In order to analyze the results we break down the variables individually to understand their relationship and how it affects the quality and diversity of the samples.
\subsection{Average IS vs Minimum segment angle}
The average IS gives a good understanding of the generated sample. We wanted to compare the average IS value when the data is shifted to mean or not. 

\begin{figure}[htpb]
\centering
\includegraphics[scale=0.4]{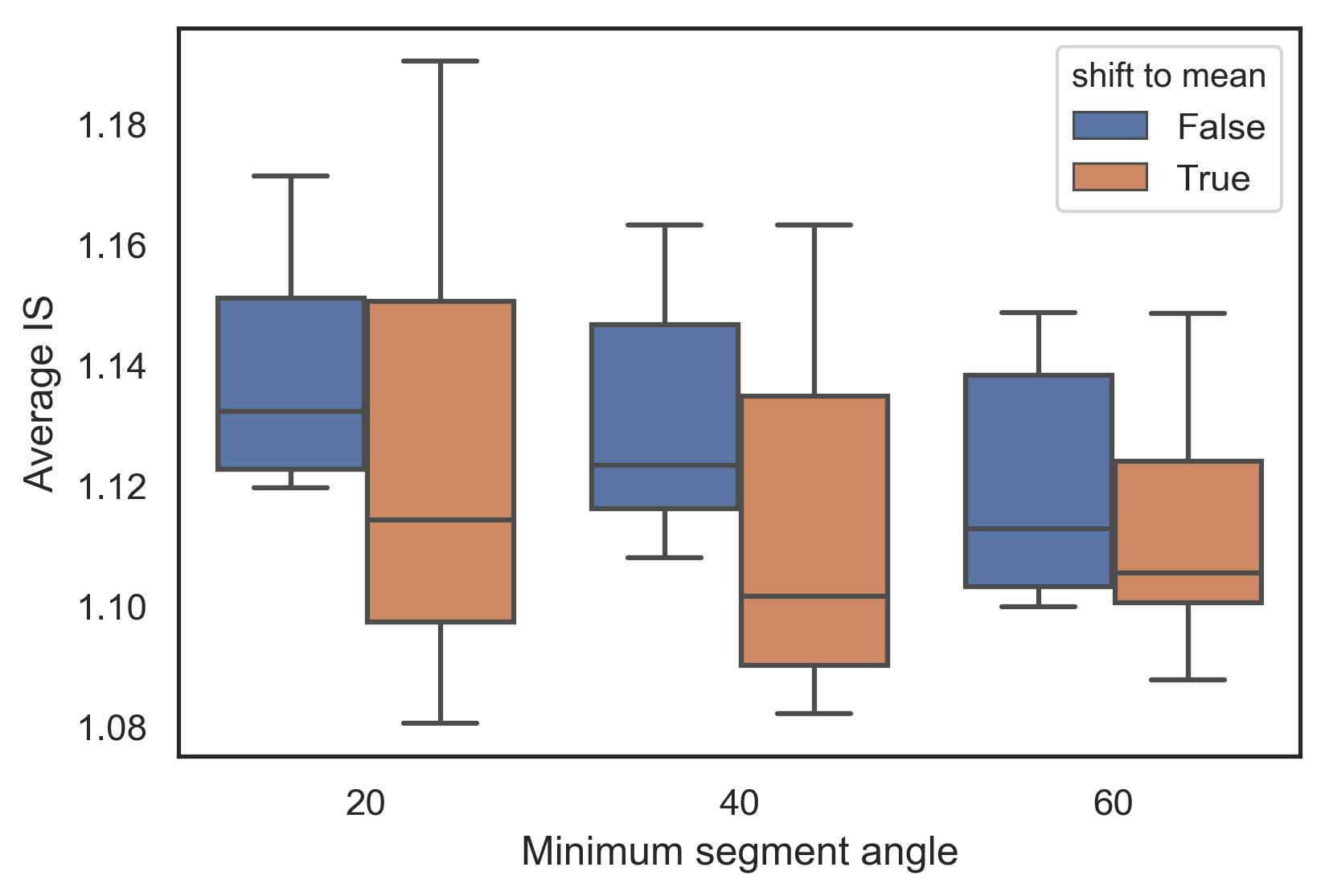}
\caption{\bf Average IS vs minimum segment angle varying shift to mean}
\label{fig:4s}
\end{figure}
We can clearly see the decreasing trend of the average IS value with increasing minimum segment angle. Also, the spread of the IS score increase if the data is shifted to mean. However the mean value decreases if the data is shifted to mean. We can see that when the data is shifted towards mean, the Inception Score decreases.  The most probable reason for this phenomenon is that shifting the data towards the mean reduces the diversity of the synthetic data and hence also the generated samples after training.


\subsection{IS standard deviation}

Standard deviation of IS requires an in-depth analysis because it helps in understanding the spread and diversity of the samples. We understand the change in standard deviation caused by:
\begin{enumerate}
    \item Shifting the data to mean.
    \item comparing it with synthetic and generated dataset.
    \item changing the number of vertices.
\end{enumerate}

\begin{figure}[htpb]
\centering
\subfigure[Shift to mean: true or false]{
\includegraphics[scale=0.4]{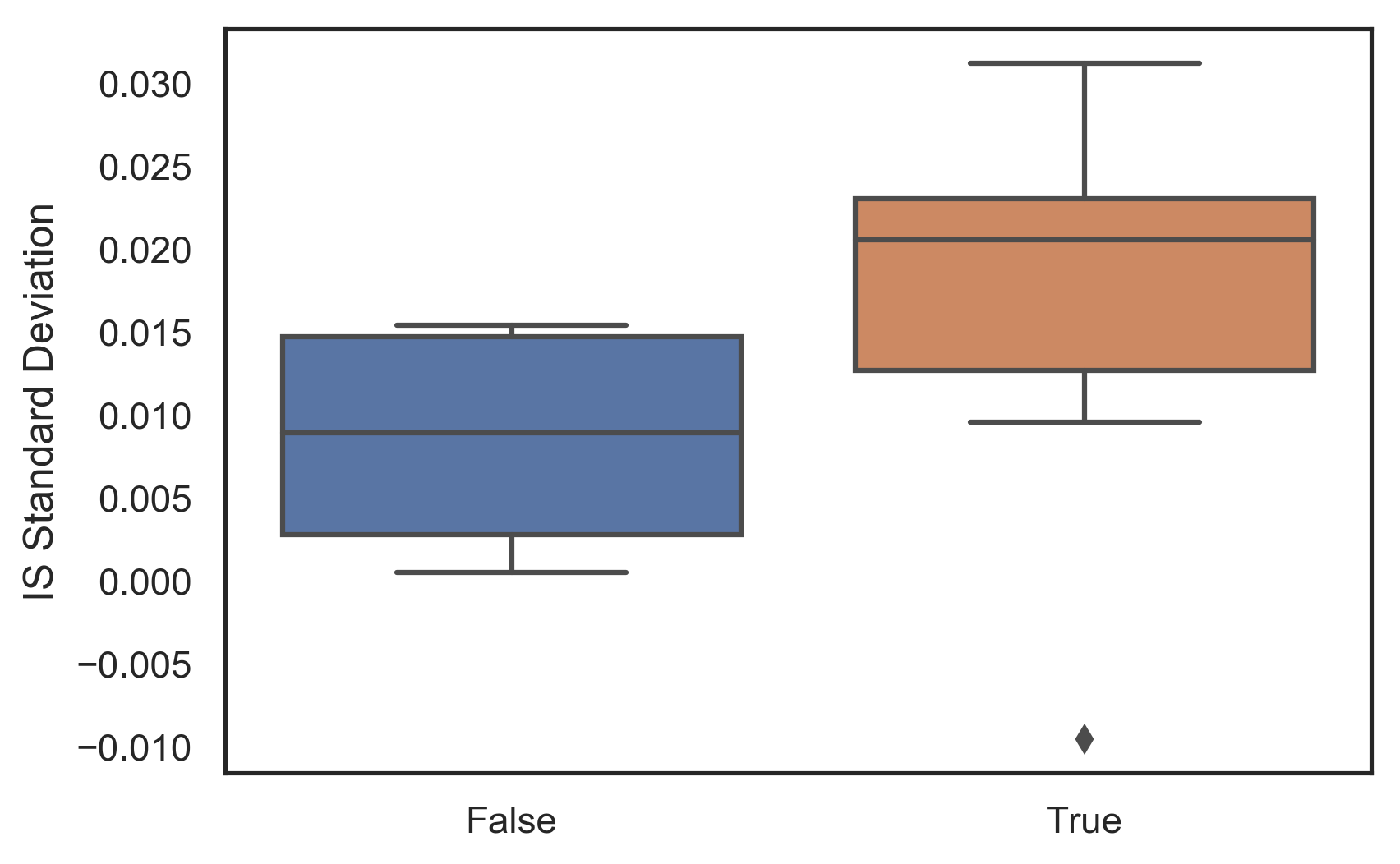}
\label{fig:1}} 
\subfigure[Generated samples]{
\includegraphics[scale=0.4]{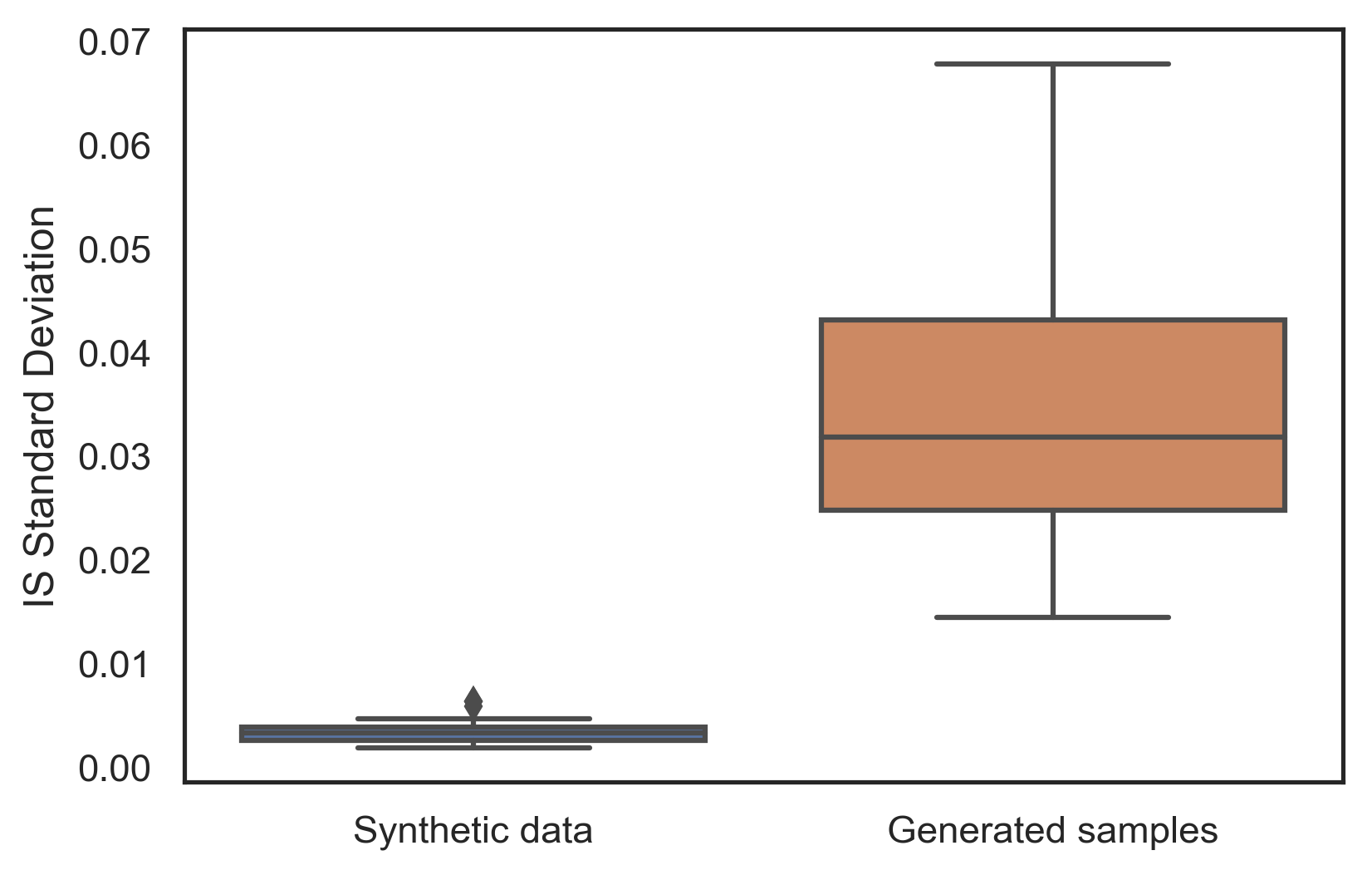}
\label{fig:2}}
\caption{\bf IS standard deviation plots}
\label{fig:4s}
\end{figure}
In figure \ref{fig:1} it can be seen that the standard deviation increase when the data is shifted to mean. The spread is also high for this case. If the data is not shifted to mean, then the standard deviation is not highly spread and is closer together.

In figure \ref{fig:2} we compare the IS standard deviation with synthetic dataset and generated samples. The standard deviation for the synthetic dataset has low values as all the samples are generated are quite regular. Whereas the ones's generated by the model have a very high standard deviation, and hence shows a large spread.

\begin{figure}[htpb]
\centering
\subfigure[Shift to mean: True or false]{
\includegraphics[scale=0.4]{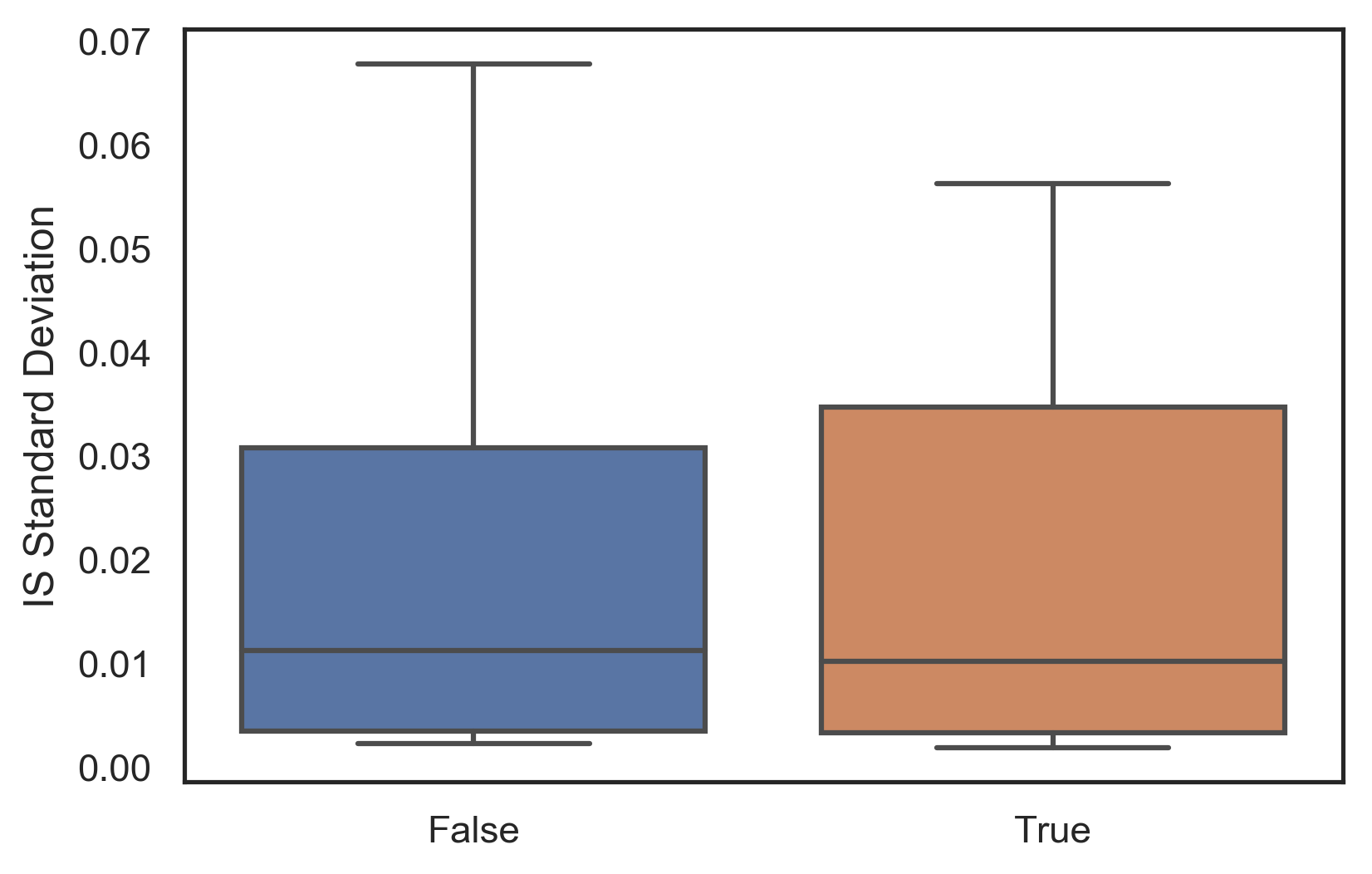}
\label{fig:3}} 
\subfigure[Number of Vertices]{
\includegraphics[scale=0.4]{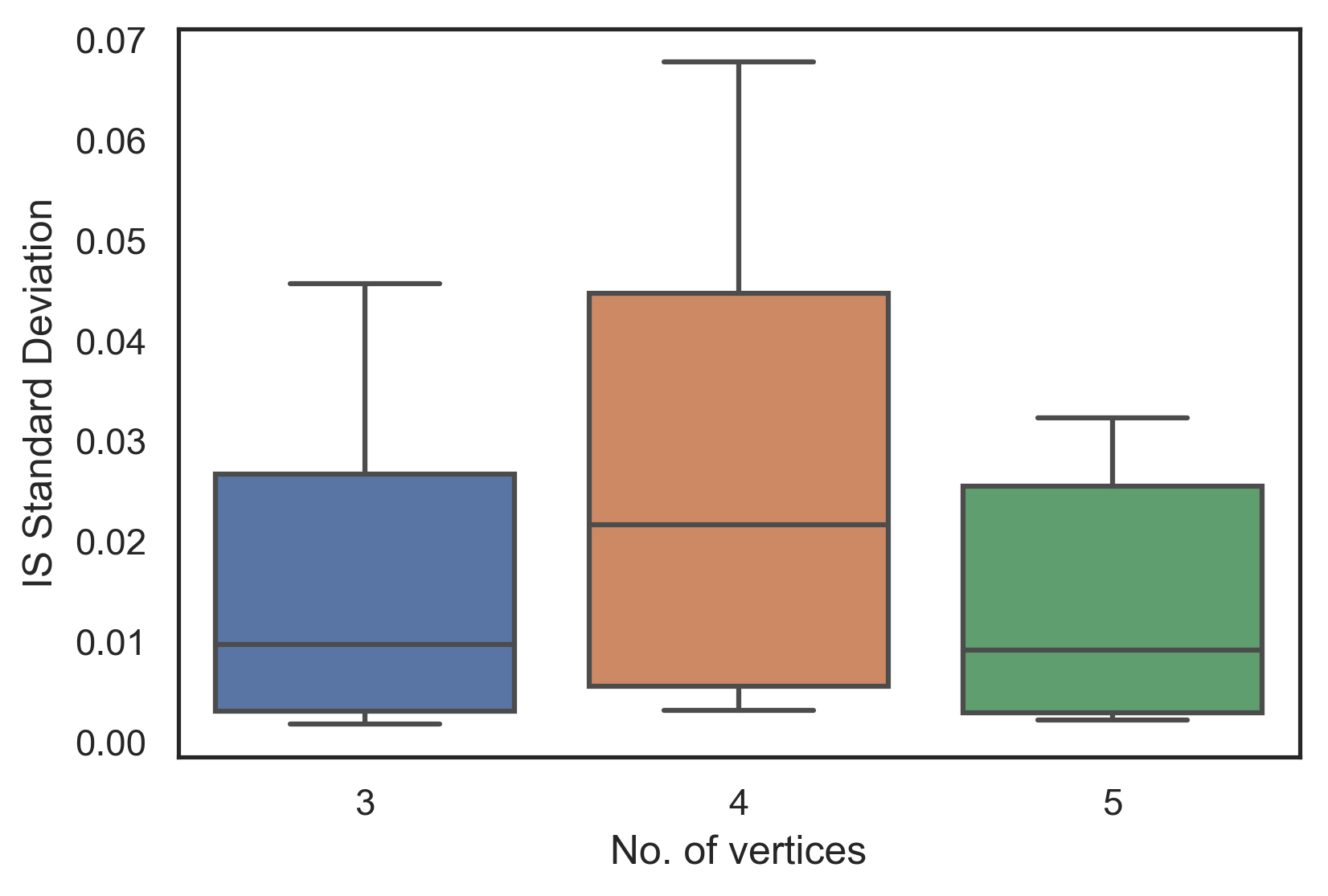}
\label{fig:4}}
\caption{\bf IS standard deviation plots}
\label{fig:4s}
\end{figure}

In figure \ref{fig:3} it can be seen that the standard deviation increase when the data is shifted to mean. The spread is also high for this case. If the data is not shifted to mean, then the standard deviation is not highly spread and is closer together.

In figure \ref{fig:4} we can see that quadrilaterals have the highest spread in terms of IS standard deviation. The pentagons have the lowest IS standard deviation because of the increasing number of sides in the polygon. However, the mean IS standard deviation is similar for both triangles and pentagons.


\subsection{Average and standard deviation of IS}
\begin{figure}[htpb]
\centering
\includegraphics[scale=0.4]{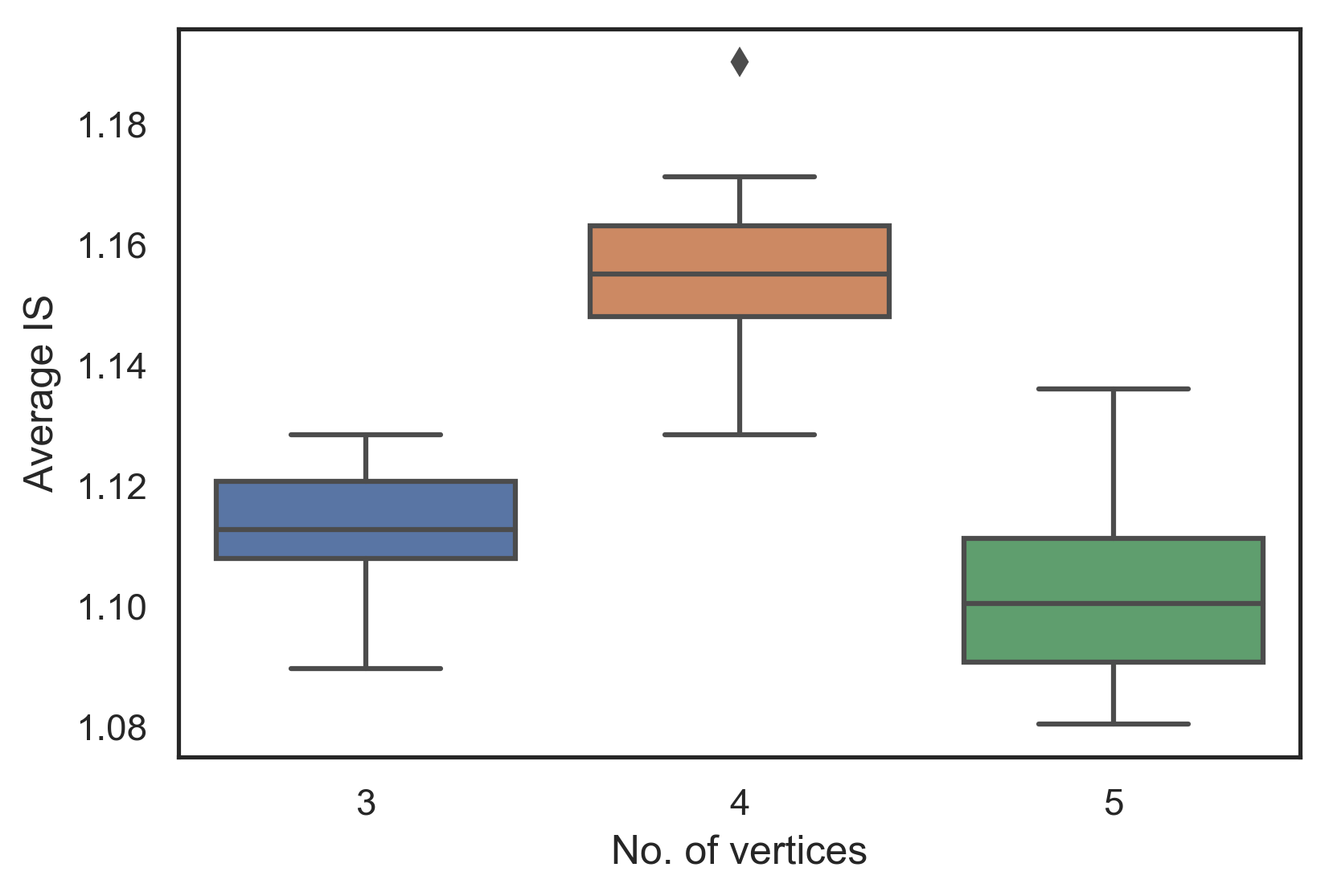}
\caption{\bf Comparison of average IS vs number of vertices}
\label{fig:avgis}
\end{figure}

According to plot \ref{fig:avgis}, quadrilaterals have the highest value of average IS. The spread for all the n-gons are similar. Also, it can be seen that pentagon has the least value of average IS. 

\begin{figure}[htpb]
\centering
\includegraphics[scale=0.4]{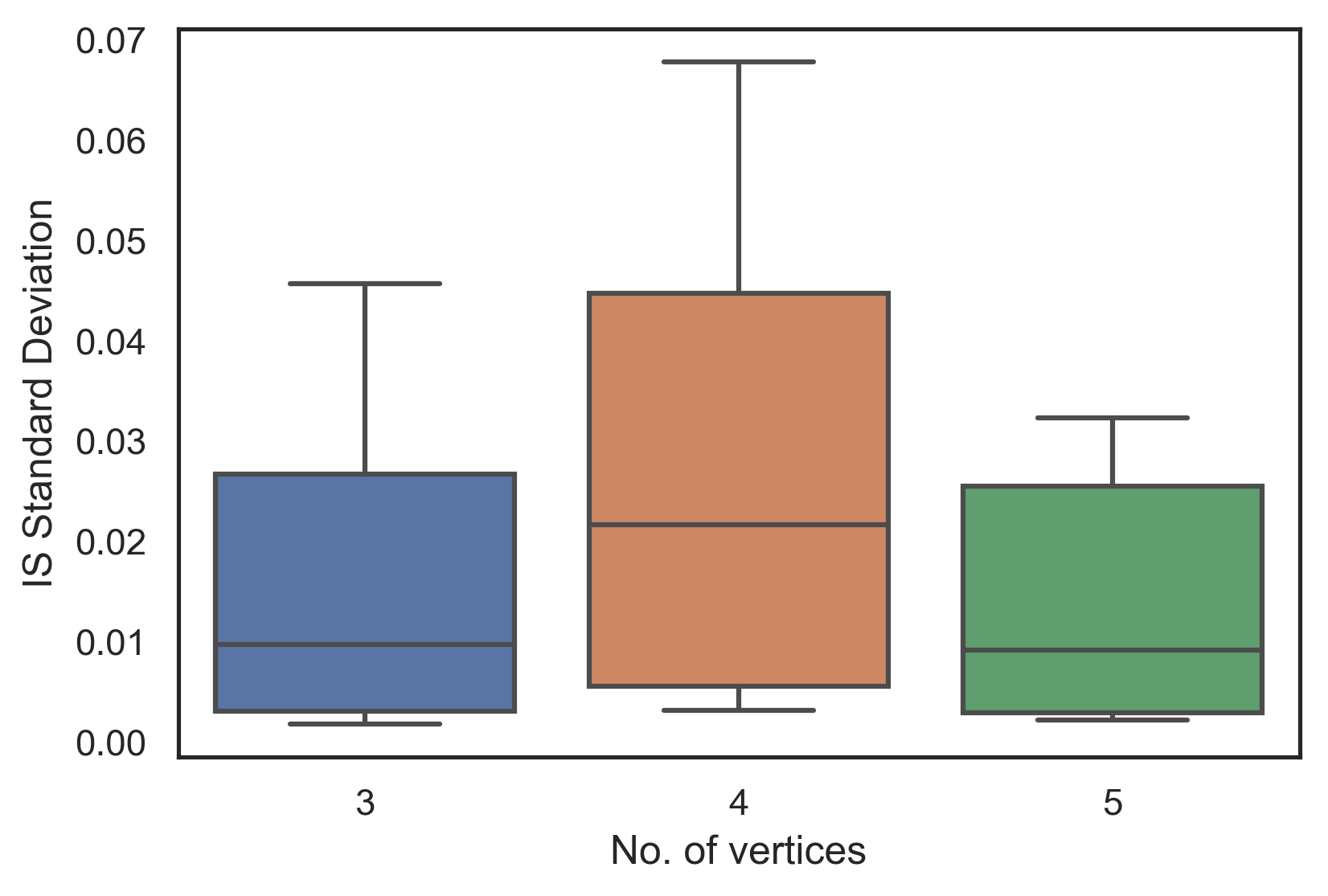}
\caption{\bf Comparison of standard deviation in IS vs number of vertices}
\label{fig:isstd}
\end{figure}

In figure \ref{fig:isstd}, the quadrilaterals have the highest spread and the pentagons have the least spread of standard deviations. The mean IS standard deviation is highest for quadrilaterals and hence it can be said that they are the most varied. 

\subsection{Difference in IS and varying vertices and minimum segment angle}
\begin{figure}[htpb]
\centering
\subfigure[Number of Vertices]{
\includegraphics[scale=0.4]{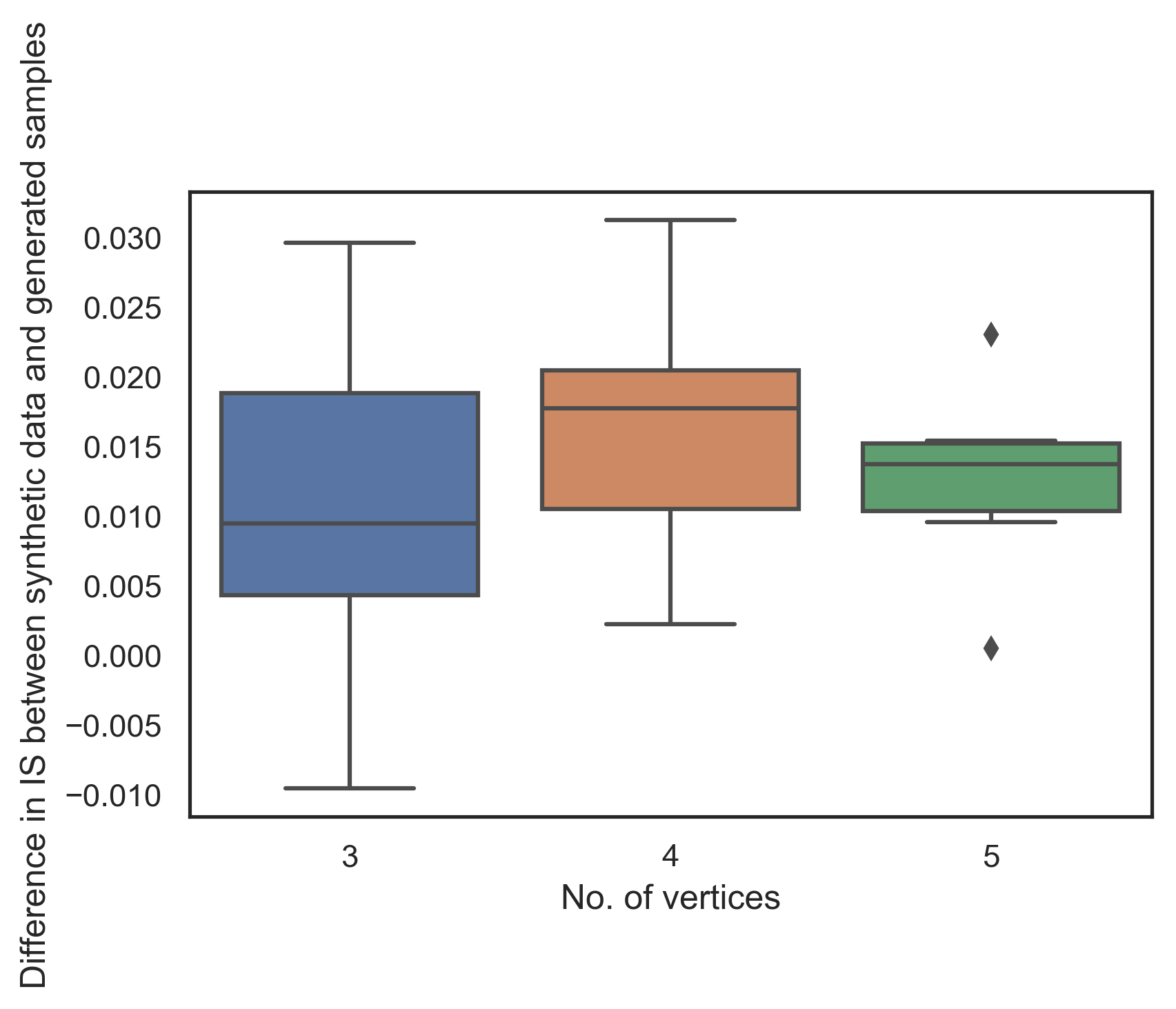}
\label{fig:vert}} 
\subfigure[Minimum segment angle]{
\includegraphics[scale=0.4]{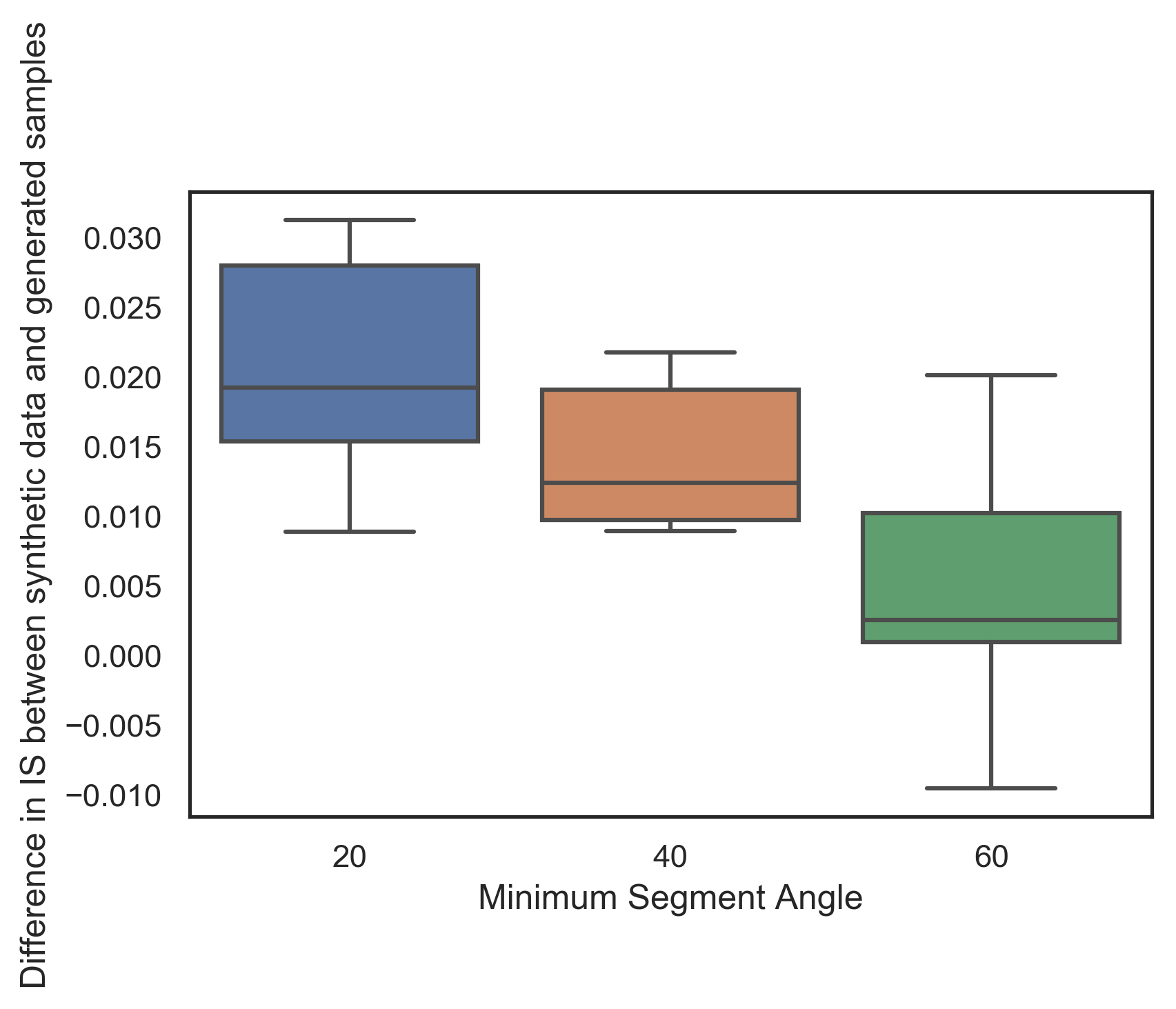}
\label{fig:msa}}
\caption{\bf Difference in IS and varying vertices and minimum segment angle}
\label{fig:4s}
\end{figure}

Varying the number of vertices, we can see in figure \ref{fig:vert} that the spread of difference of IS [between synthetic data and generated samples] increases as the number of vertices increases. This means that the images generated with higher vertices has the quality similar to that of the input sample image. Whereas for lower vertices, it can be seen that the quality of the generated image may be better that the sample image.

However, in figure \ref{fig:msa} we can see that the difference keeps on decreasing as the minimum segment angle decreases. This means that for some generated samples the IS score is better than the input sample.

\subsection{Comparing individual and combined dataset}
\begin{figure}[htpb]
\centering
\includegraphics[scale=0.4]{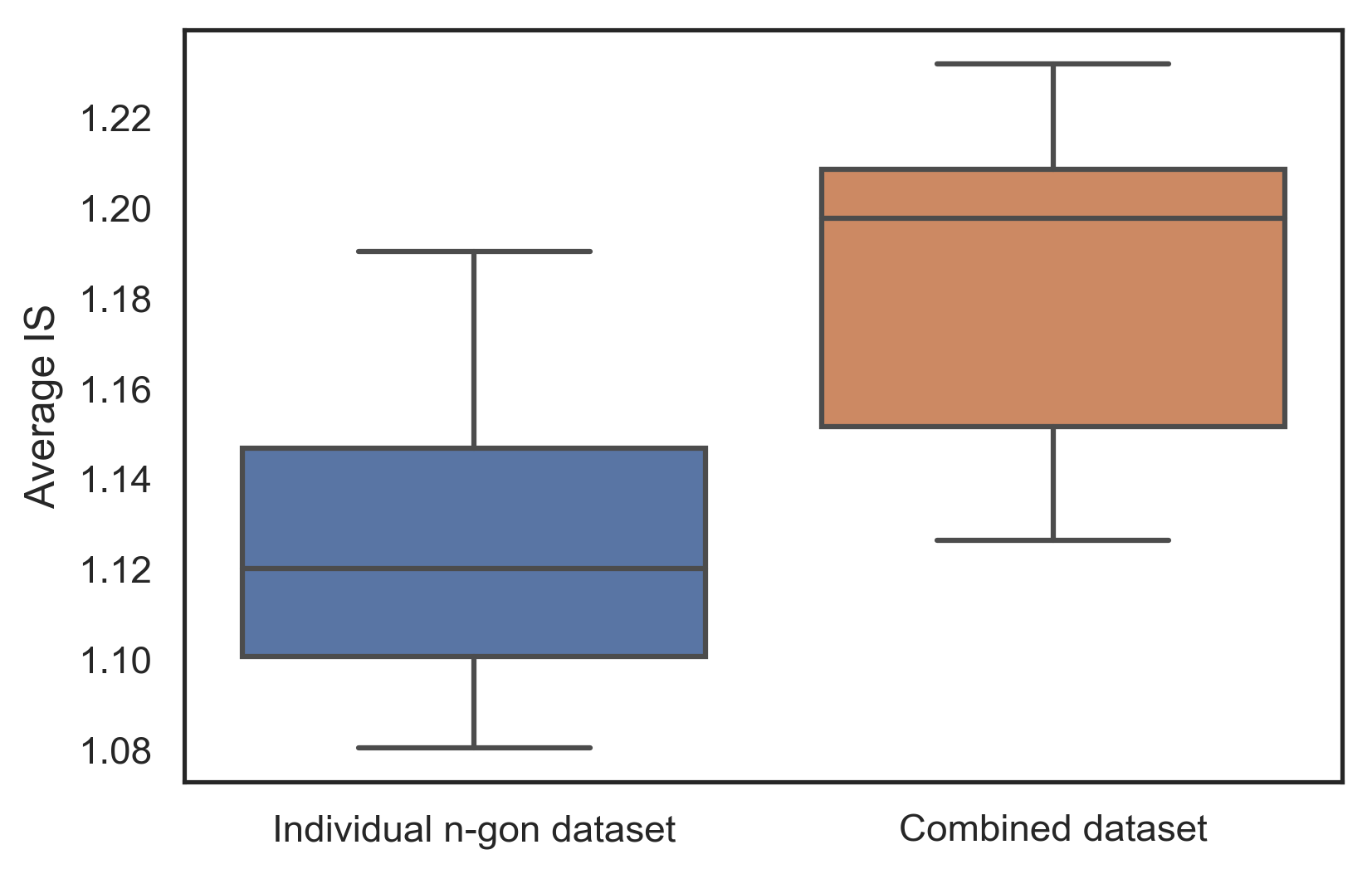}
\caption{\bf Comparison of average IS vs number of vertices}
\label{fig:avgis1}
\end{figure}

\begin{figure}[htpb]
\centering
\includegraphics[scale=0.4]{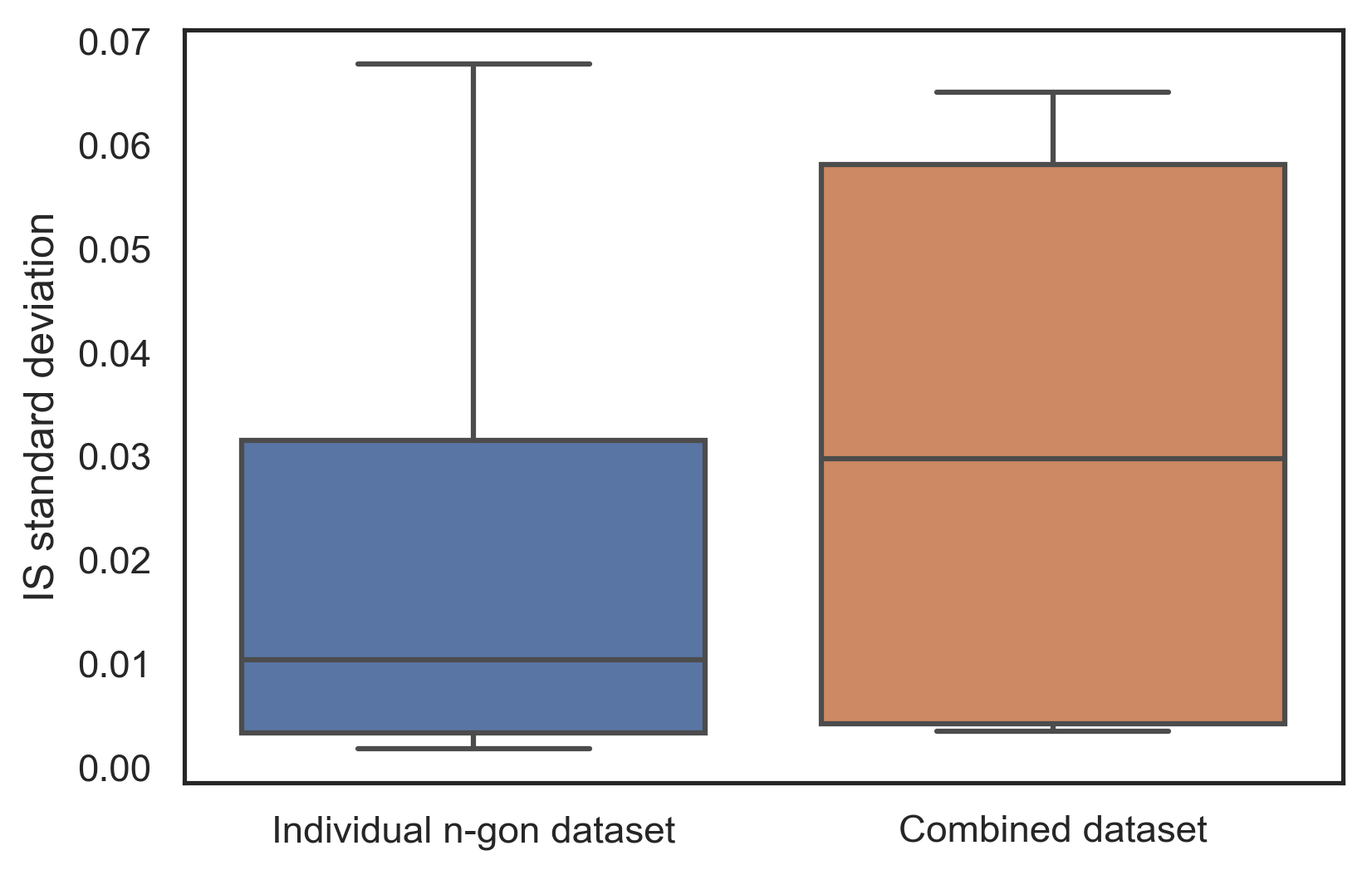}
\caption{\bf Comparison of standard deviation in IS vs number of vertices}
\label{fig:isstd1}
\end{figure}

According to plot \ref{fig:avgis1}, Combined dataset containing all the n-gons have higher average IS whereas individual n-gons have a wider spread of IS values. In figure \ref{fig:isstd1}, although the combined dataset has higher standard deviation, the maximum standard deviation for both are same. The mean standard deviation for n-gons data set is less than that for combined dataset. 

\subsection{Difference in IS after shifting towards mean}
An important criteria to check is the difference in IS when the dataset is shifted towards mean. We can clearly see that for quadrilaterals, the difference increases in both the synthetic as well as generated data. 

\begin{figure}[htpb]
\centering
\subfigure[Synthetic data]{
\includegraphics[scale=0.4]{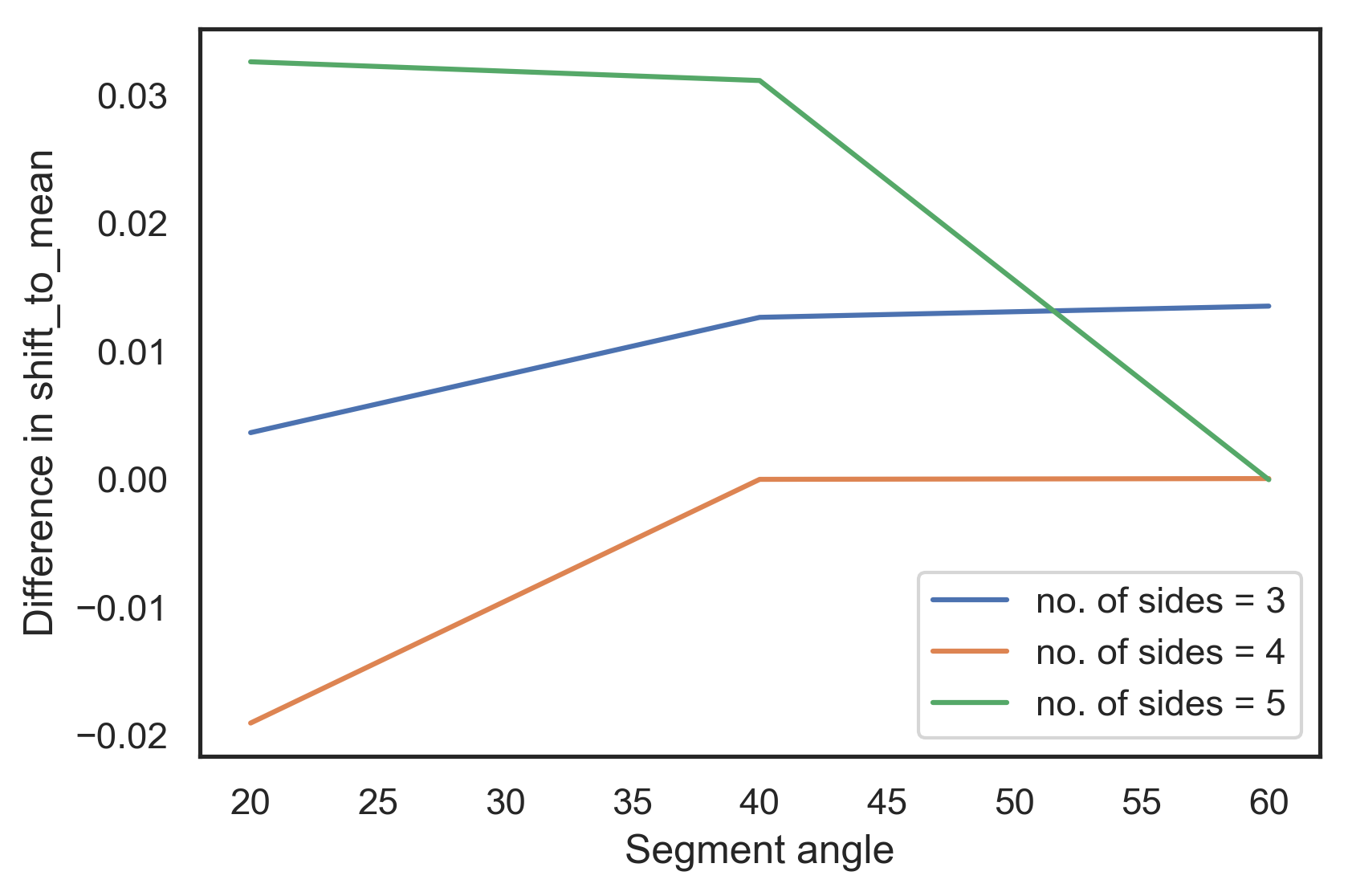}
\label{fig:ellipse}} 
\subfigure[Generated samples]{
\includegraphics[scale=0.4]{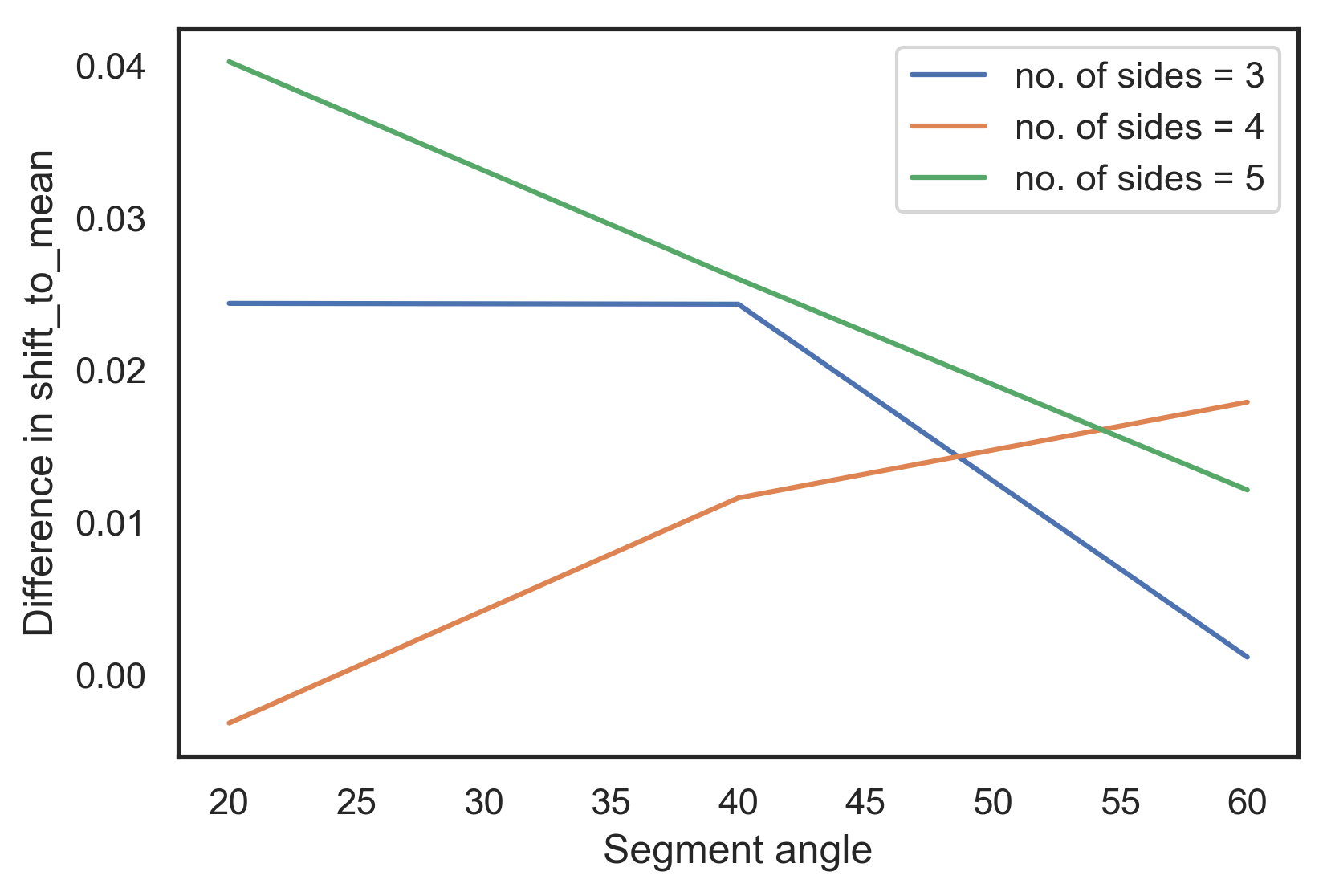}
\label{fig:scotch}}
\caption{\bf Plot Between difference in shift to mean vs segment angle for synthetic and result data}
\label{fig:4s}
\end{figure}

Whereas for triangles, the difference decreases for both the cases if the segement angle increase. however in quadrilaterals, the trend changes if the segment angle is more than 40 degrees.


\subsection{Difference in average IS between synthetic data and generated samples}
If the data is not shifted to mean, the difference between the values of average IS for synthetic data and generated samples decreases for all the n-gons. However if the data is shifted to mean, then the difference decrease only of the shapes which has minimum segment angle less than 40 degrees. If the minimum segment angle is increase, then the trend is increasing for quadrilaterals and pentagons whereas steep decrease for triangles.
\begin{figure}[htpb]
\centering
\subfigure[Shift to mean: True]{
\includegraphics[scale=0.4]{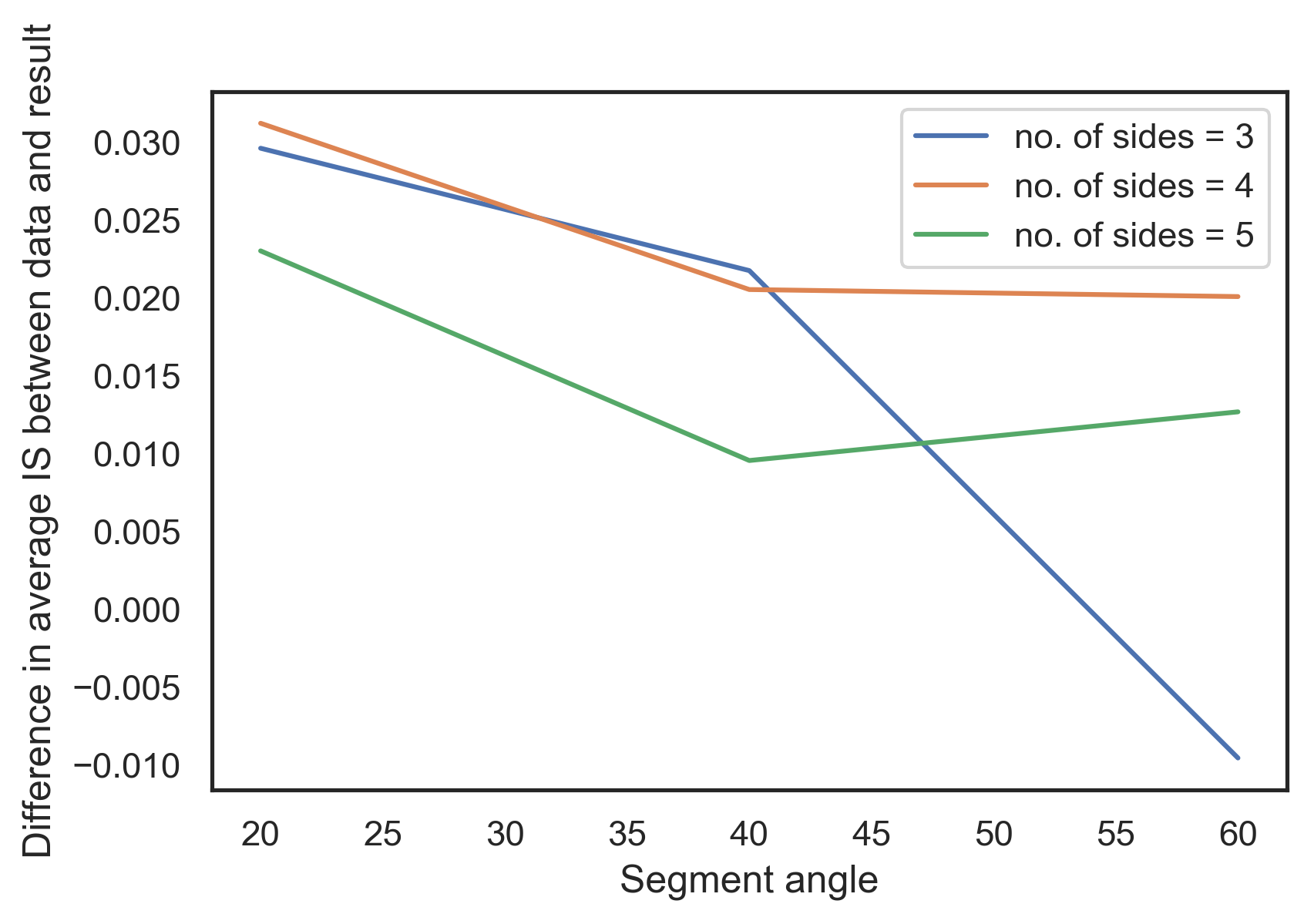}
\label{fig:ellipse}} 
\subfigure[Shift to mean:: False]{
\includegraphics[scale=0.4]{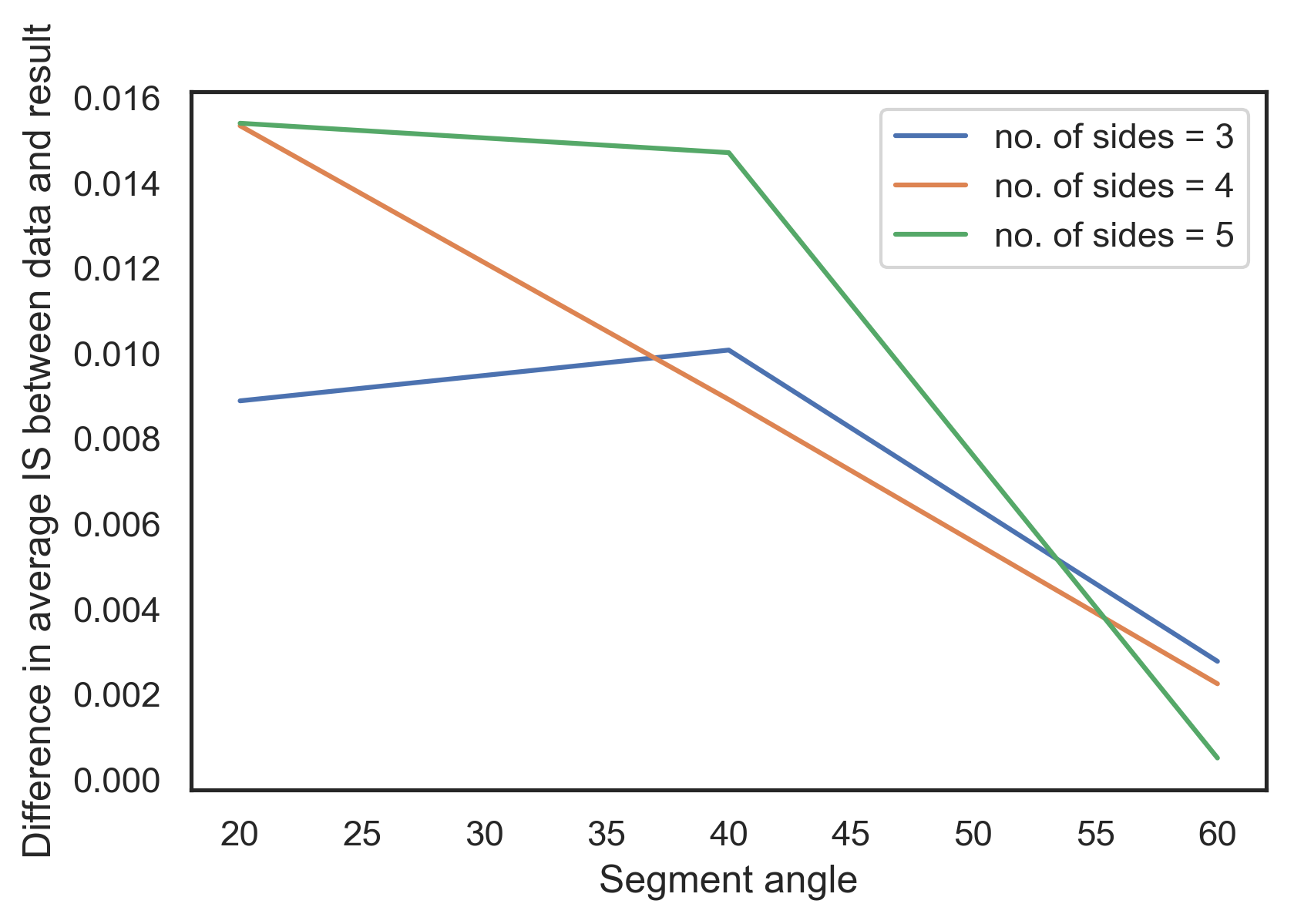}
\label{fig:scotch}}
\caption{\bf Difference in average IS between synthetic data and generated samples and varying shift to mean}
\label{fig:4s}
\end{figure}


\section{Analysing the loss surface}

\begin{figure}[htpb]
\centering
\includegraphics[scale=0.5]{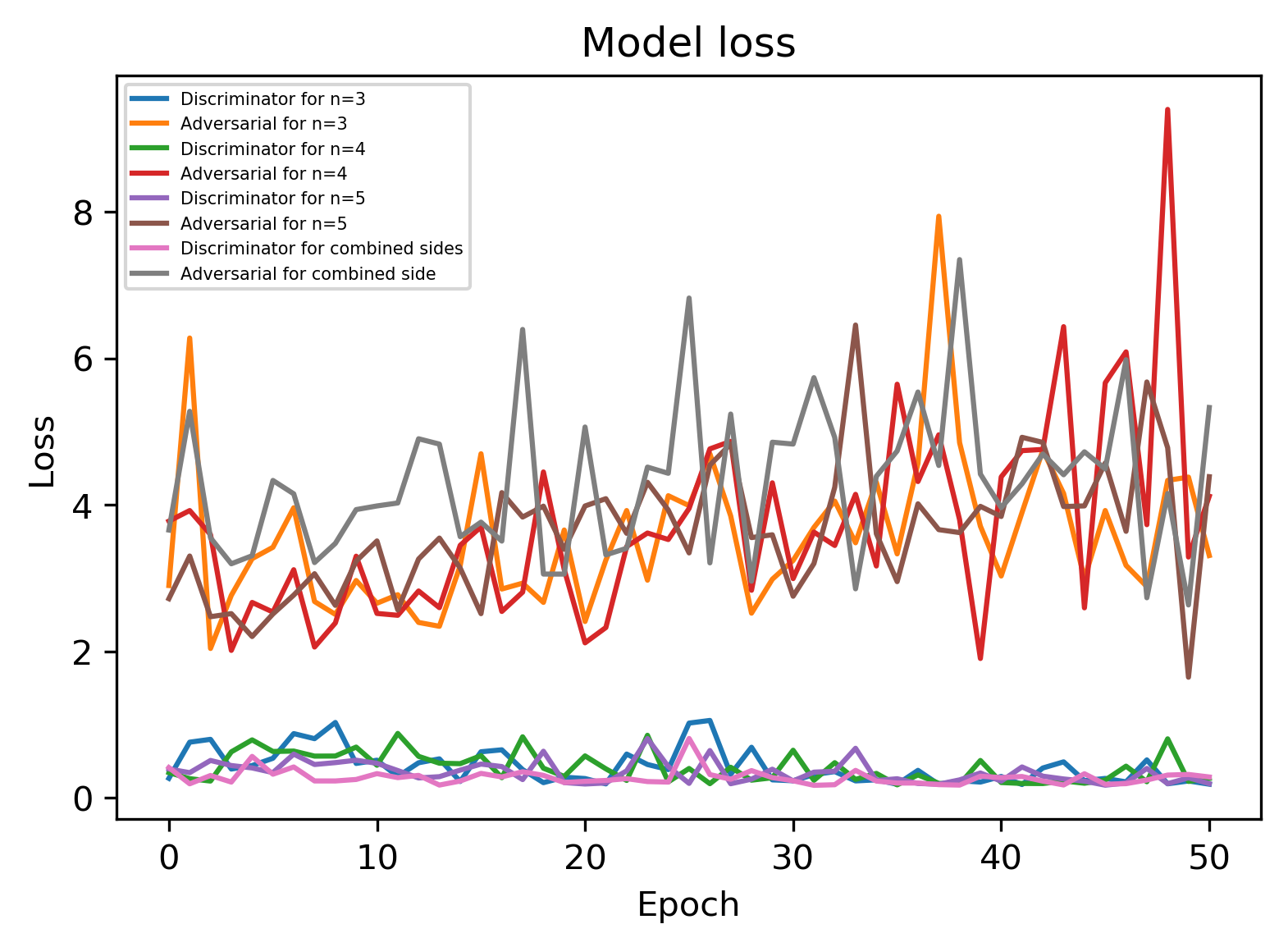}
\caption{\bf Model loss}
\label{fig:loss1}
\end{figure}

To understand the training process of the GAN model, we plotted the model loss, generator loss and discriminator loss. Although IC increases as the number of classes increase, the training of GANs become more difficult since the loss landscape of the generator becomes less smoother. In figure \ref{fig:loss1} we can see that the discriminator loss for the combined dataset is fairly smooth and hence the discriminator is performing good. Although the loss curve for the adversarial of the combined dataset is not very smooth, it performs better than the rest of the adversarials.

\begin{figure}[htpb]
\centering
\subfigure[Generator Loss]{
\includegraphics[scale=0.4]{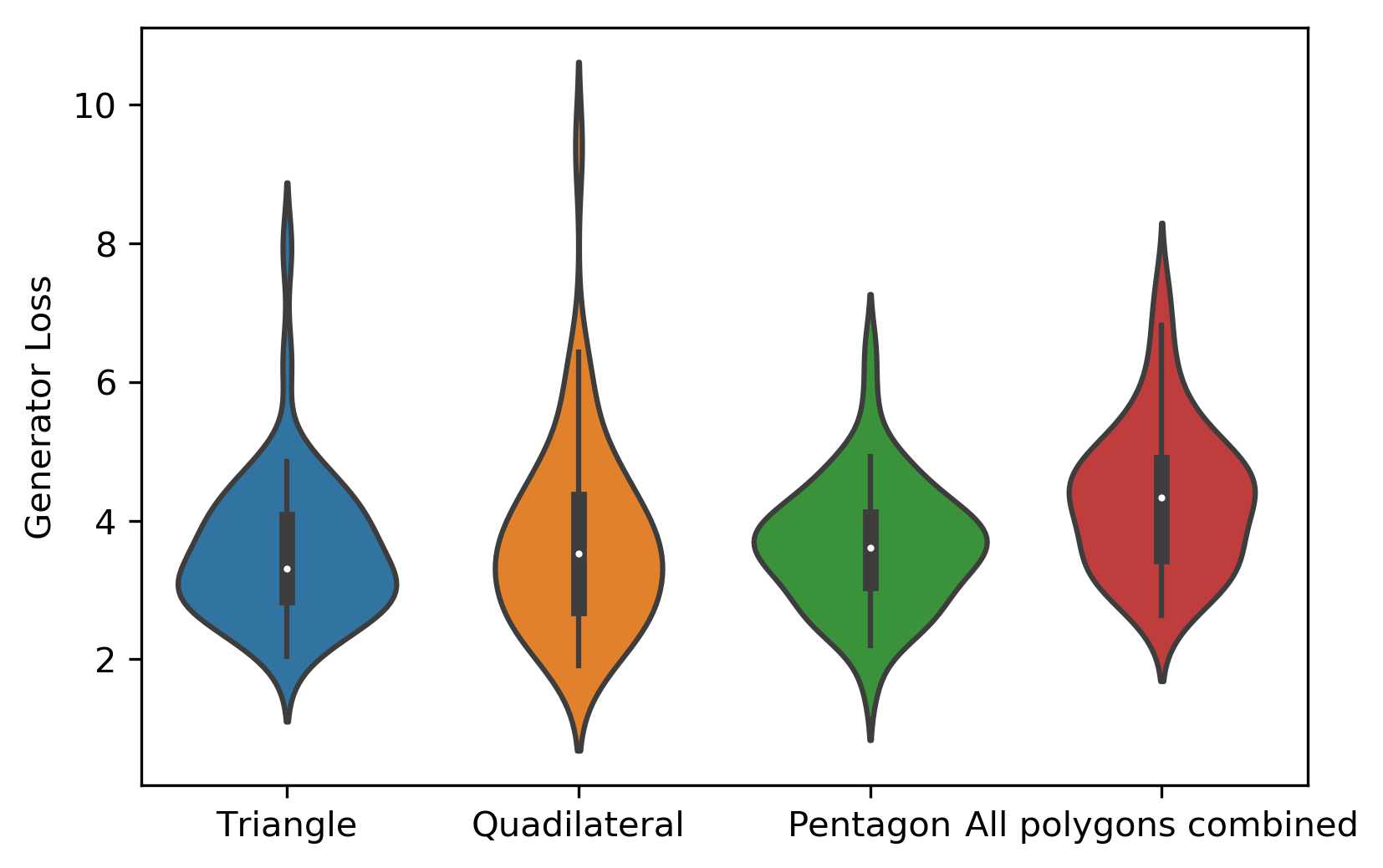}
\label{fig:vgis1}} 
\subfigure[Discriminator Loss]{
\includegraphics[scale=0.4]{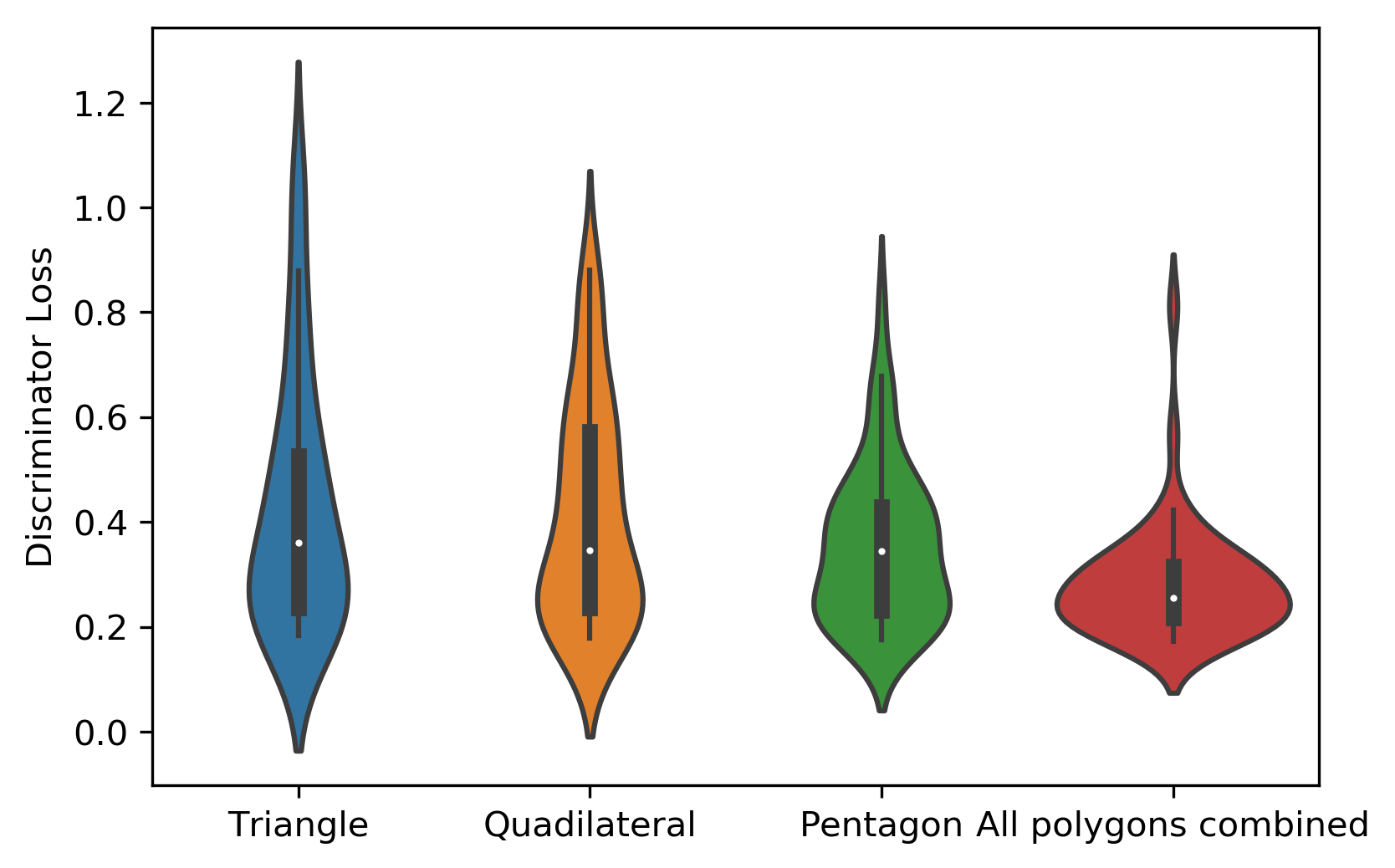}
\label{fig:istd1}}
\caption{\bf Losses}
\label{fig:loss}
\end{figure}

When the number of classes increase the discriminator loss landscape is smoother  and  achieves  convergence  faster  but  the  generator  loss  landscape  becomes  less smoother and the loss keeps fluctuating since the model parameters oscillate. In figure \ref{fig:loss} we can see that the model parameters of the generator keep oscillating as the number of classes increase which destabilizes the training process and thus, convergence becomes harder and slower. When all the claases are trained together, the discriminator loss plot is dense while the generator loss is higher. Moreover, this training imbalance between the generator and the discriminator causes overfitting and may also lead to mode collapse.

\section{Conclusion}

Training GANs can be a notoriously difficult challenge and it is even harder to train on certain datasets. In this project we explored characteristic traits of datasets which affect training of GANs. We empirically proved which characteristics make training more stable and lead to generation of better quality samples. 

During our study we found the following traits affect training of GANs and Inception Score which depends on the quality and diversity of the generated samples:

1. Shifting the data towards mean : In our experiments we found that when the data is shifted towards mean, the Inception Score decreases. The most probable reason for this phenomenon is that shifting the data towards the mean reduces the diversity of the synthetic data and hence also the generated samples after training. Moreover, we did not see any noticeable stability during the training phase when the dataset is shifted towards the mean.

2. Regularity : We empirically confirm that it is when the data is more regular, the training of the GAN is more smoother as confirmed by the smoother loss lanscapes. Moreover, quadilaterals had a higher IS than 3 and 5 sided polygons owing to their more regular geometric structure. We also noticed that as the number of vertices increased IS decreased.

3. Minimum segment angle : As the minimum segment angle increased, IS decreased. This can be attributed to the fact that as minimum segment angle increases, diversity in the synthetic data decreases. By increasing the minimum segment angle we restrict possible the ways to construct a polygon, thereby decreasing diversity and since IS increases with increase in diversity, a lower minimum segment angle increases IS. 

4. Diversity : As the number of classes increase, we see an increase in IS since the diversity of the generated samples increase. 

5. Number of classes : Although IC increases as the number of classes increase, the training of GANs become more difficult since the loss landscape of the generator becomes less smoother. We found that on increasing the number of classes the discriminator loss landscape is smoother and achieves convergence faster but the generator loss landscape becomes less smoother and the loss keeps fluctuating since the model parameters oscillate. GANs work on the principle of a zero-sum non-cooperative game where the model convergence is dependent on the generator and the discriminator reaching Nash equilibrium. Since the model parameters of the generator keep oscillating as the number of classes increase, the training destabilizes and convergence becomes harder and slower. Moreover, this training imbalance between the generator and the discriminator causes overfitting and may also lead to mode collapse.

Through this study we found out which traits affect learnability and tried to explain these phenomenons emperically. By identifying these traits we can particularly tackle key issues affecting learnability and prepare and process data such that the training landscape becomes smoother and more stable, and thus yielding better quality samples. This can be extremely useful to perform better architecture search especially in Generative Teaching Networks (GTNs) \cite{such2019generative} which generate synthetic data that can help other neural networks to rapidly learn the distribution when trained on real data.

\section{Future Enhancement}
More n-sided polygons with diverse statistical properties can be explored to further cement our conclusion and draw more insights about the properties which stabilize the training of GANs.

We use Inception Score to compare the quality of the generated samples. We found a strong correlation between average IS and standard deviation in IS (~0.77) but we could not find a partcular reason for this phenomenon. We think that an explanation for this phenomenon would give us a better understanding of IS itself. Other GAN evaluation metrics like Frechet Inception Distance (FID), MS-SSIM which can seperately evaluate the diversity of the generated samples, Geometric Score and precision and recall of GANs can be further investigated by evaluating it on our generated samples. It might be even interesting to explore how these metrics compare to each other and if they show good correlation. While a human judge would be the best evaluator to rate the quality of generated samples, it is both expensive and subjective. A new comprehensive GAN evaluation metric that is not overly influenced by a single factor would be a major breakthrough in GAN evaluation as it would allow us to compare various GAN architectures and assess which architectures are better suited for a particular task. 

Here we found that distributions which are more regular are easier for a GAN to model, it is worth exploring how we can overcome this. A latent representation which is more regular would lead to a smoother training landscape and thus yield better quality samples.


\end{document}